\documentclass{article}

\usepackage[preprint]{neurips_2021}

\usepackage[utf8]{inputenc} % allow utf-8 input
\usepackage[T1]{fontenc}    % use 8-bit T1 fonts
\usepackage{url}            % simple URL typesetting
\usepackage{booktabs}       % professional-quality tables
\usepackage{amsfonts}       % blackboard math symbols
\usepackage{nicefrac}       % compact symbols for 1/2, etc.
\usepackage{microtype}      % microtypography
\usepackage{amsmath}
\usepackage{subcaption}
\usepackage{algorithm}
\usepackage{algorithmic}
\usepackage{graphicx}
\usepackage{bm}
\usepackage{enumitem}

\usepackage{multirow}
\usepackage[table,x11names,dvipsnames,table]{xcolor}

\usepackage[numbers]{natbib}

\title{Scalable Whitebox Attacks on Tree-based Models}

\author{%
  Giuseppe Castiglione\thanks{Authors contributed equally to this work.} \\
  Borealis AI\\
   \And
   Gavin Ding\footnotemark[1] \\
   Borealis AI \\
   \AND
   Masoud Hashemi\footnotemark[1] \\
   Borealis AI \\
   \And
   Christopher Srinivasa \\
   Borealis AI \\
   \And
   Ga Wu\thanks{Main contact of this work, email: ga.wu@borealisai.com} \\
   Borealis AI \\
}

\begin{document}

\maketitle

\begin{abstract}
Adversarial robustness is one of the essential safety criteria for guaranteeing the reliability of machine learning models. While various adversarial robustness testing approaches were introduced in the last decade, we note that most of them are incompatible with non-differentiable models such as tree ensembles. Since tree ensembles are widely used in industry, this reveals a crucial gap between adversarial robustness research and practical applications. This paper proposes a novel whitebox adversarial robustness testing approach for tree ensemble models. Concretely, the proposed approach \textit{smooths} the tree ensembles through temperature-controlled sigmoid functions, which enables gradient descent-based adversarial attacks. By leveraging sampling and the \textit{log-derivative} trick, the proposed approach can scale up to testing tasks that were previously unmanageable. We compare the approach against both random perturbations and blackbox approaches on multiple public datasets (and corresponding models). Our results show that the proposed method can 1) successfully reveal the adversarial vulnerability of tree ensemble models without causing computational pressure for testing and 2) flexibly balance the search performance and time complexity to meet various testing criteria.
\end{abstract}

% Warning: smoothing tree is not our contribution, but we are the one use the smoothed tree to facilitate adversarial testing.

\section{Introduction}
Machine learning~(ML) models are proven to be vulnerable to adversarial examples~\cite{fawzi2018adversarial,carlini2019evaluating}; small but carefully crafted distortions of inputs created by adversaries could fool the ML models. For safety-sensitive applications (e.g. finance~\cite{dixon2020machine}, health service~\cite{jiang2017artificial}, and autopilot systems~\cite{morgulis2019fooling}), such an issue could result in catastrophic consequences.
This has attracted much attention from the research community~(see \cite{chakraborty2018adversarial,cohen2019certified}). In particular, effectively testing the adversarial robustness before model deployment is one of the most crucial challenges~\cite{carlini2019evaluating}. 

In the literature, many adversarial robustness testing approaches mainly focus on deep learning models. 
For instance, \cite{goodfellow2014explaining} proposed Fast Gradient Sign Method~(FGSM) to test the robustness of neural networks by generating adversarial examples through a one-step update.
Later, the Basic Iterative Method~(BIM)~\cite{kurakin2016adversarial} extended FGSM by introducing multi-step gradient updates that result in a better success rate.
%\cite{katz2017reluplex} proposed SMT Solver based model verification method for neural network models with Relu activation function. \chris{this is verification, are you sure you want to include it in the same section as testing?}
\cite{gopinath2017deepsafe} introduced a data-guided methodology to determine regions that are likely to be safe (instead of focusing on individual points). 
%And, to demonstrate the testing effectiveness, most of the proposed testing approaches are evaluated on image tasks that are hard to verify in other application domains ~(see~\cite{dong2020benchmarking,zeng2019adversarial}). 

Despite their testing effectiveness,  we note that few of the existing approaches support non-differentiable models that are widely used in the industry. Indeed, many ML models used in product lines are tree ensemble models~(see \cite{ahmad2017trees,rasheed2021explainable,gade2019explainable}) due to model computational efficiency and transparency (with readable interpretations for humans). This fact reveals a crucial gap between adversarial robustness research and practical applications. To address this research gap, 
\cite{kantchelian2016evasion} proposed a white-box attack method using Mixed Integer Linear Programming~(MILP) to avoid computing gradients.
\cite{chen2019robust}, alternatively, formulated the adversarial robustness test into a maximum clique enumeration task which shows better scalability than the MILP approach.
While these approaches are enlightening, they appear to be computationally intractable in practice, where the testing task may involve an ensemble with hundreds of trees, thousands of features, and millions of data points to test on. Hence, blackbox attack approaches~\cite{ilyas2018black,alzantot2019genattack} become a relatively practical option for large-scale testing, while still computationally expensive. 

In this paper, we propose a scalable whitebox adversarial robustness testing approach for tree ensemble models. In particular, we aim to unlock gradient descent-based adversarial robustness testing on tree ensemble models by smoothing the trees. By replacing each decision node in a tree with a temperature-controlled sigmoid function, we can approximate the original target model with a controllable error gap. To facilitate efficient adversarial example search, we propose two approaches by either injecting sparse noise during gradient descent or introducing a log-derivative trick and Monte Carlo sampling to approximate the gradients. Both variants provide sufficient stochasticity that maximize the search coverage. 

To demonstrate the effectiveness of the proposed method, we compared it against multiple approaches, such as random perturbations, and black box attack methods on multiple public datasets and corresponding models. Our experimental results show the proposed method can 1) successfully reveal the adversarial vulnerability of tree ensembles without causing computational pressure for testing, and 2) flexibly balance the search performance and time complexity to meet various testing criteria.  
\vspace{-2mm}
\section{Preliminary}
\label{sec:preliminary}
\vspace{-2mm}
%\subsection{Notations}
In this section, to ground our main contribution, we highlight previous research in the field of adversarial robustness testing. We then review existing work on ensembles of decision trees to facilitate our description of the proposed attack.% To  facilitate  our description of the proposed method, we will also review the tree ensemble models
\vspace{-2mm}
\subsection{Adversarial Robustness Testing}
\vspace{-2mm}
The vulnerability of deep networks to adversarial attacks was first introduced in \cite{szegedy2013intriguing}, which demonstrated that small, imperceptible perturbations to a model's inputs were capable of dramatically changing its outputs.
% \cite{goodfellow2014explaining} {\color{green} introduced} the Fast Gradient Sign Method~(FGSM) to test the robustness of neural networks by generating adversarial examples through a one-step update. Later, the Basic Iterative Method~(BIM)~\cite{kurakin2016adversarial} extended FGSM by using multi-step gradient updates, resulting in a stronger attack with a higher success rate.
Aside from the iconic adversarial attack approaches, such as FGSM~\cite{goodfellow2014explaining} and BIM~\cite{kurakin2016adversarial}, mentioned in the introduction section, there are many extensions to improve the adversarial example searching efficiency.
%The issue of adversarial robustness was first introduced by \cite{szegedy2013intriguing} in 2014, which revealed the vulnerability of deep networks to adversarial noise. \cite{goodfellow2014explaining}, then proposed the Fast Gradient Sign Method~(FGSM) to test the robustness of neural networks by generating adversarial examples through a one-step update. Later, the Basic Iterative Method~(BIM)~\cite{kurakin2016adversarial} extended FGSM by introducing multi-step gradient updates that result in a better success rate. 
% \chris{The 2 previous sentences also appear exactly as is in the introduction. You should decide which of the 2 places is appropriate to avoid repetition.}
% Similar to the BIM, 
For example, Projected Gradient Descent~(PGD)~\cite{madry2017towards} suggested random initialization for adversarial example search. And, DeepFool~\cite{moosavi2016deepfool}, on another hand, focuses on generating adversarial examples by minimizing perturbations.These approaches all formulate an adversarial objective, which aims to minimize classification performance while using some norm to bound the perturbation.  These perturbations are then generated using gradient descent, which is computed by backpropagating through the target model.  These approaches thus make two assumptions: (1) the model is fully accessible, and thus these are classified as white-box attacks. (2) the model is differentiable.  %All of the above approaches are based on the assumption that the target model is accessible during adversarial example search, also called white-box attacks in the literature.

When either of these assumptions are violated, we must consider different attack algorithms.  A common strategy is to estimate gradient information through numerical approximation, as done using finite difference methods in ZOO~\cite{chen2017zoo}, or by drawing random samples (as in NES~\cite{ilyas2018black} and SPSA~\cite{uesato2018adversarial}).  Alternatively, gradient-free optimizers may be employed, such as genetic algorithms in GenAttack~\citep{alzantot2019genattack}.  In addition to their effectiveness in constructing adversarial examples, these algorithms may be ranked according to the number of queries needed to run the attack.

%If we consider a more practical situation where the model is inaccessible or non-differentiable, many works focus on estimating gradient information through numerical approximation. ZOO~\cite{chen2017zoo} estimates the gradient at each coordinate by finite differences. Similarly, NES~\cite{ilyas2018black} and SPSA~\cite{uesato2018adversarial} numerically estimate gradients by drawing random samples. Later research~(see \cite{tramer2017space,fawzi2018adversarial,naseer2019cross,xie2019improving,demontis2019adversarial}) demonstrated that adversarial examples could be transferred across models; the adversarial example crafted on one model is likely to fool other models. This discovery reflects another line of black-box adversarial robustness testing by training duplicated models for adversarial example search~\cite{liu2016delving}. 
%\chris{maybe expand a bit on how this last one works? one sentence}

\vspace{-2mm}
\subsection{Decision Trees and Tree Ensemble Models}
\label{sec:tree_ensemble}
\vspace{-2mm}
%\chris{why do you have soft trees in the section title if we do not talk about soft trees in this section?}
{\bf A Decision Tree} makes a prediction based on the value of the leaf node to which the input observation $\mathbf{x}$ gets routed, where the leaf node is determined by following the decision trajectory of the input $\mathbf{x}$ from the root node in the tree. Specifically, in each internal decision node $k$ of a tree $t$ denoted as $\mathcal{D}_{t,k}$, the input $\mathbf{x}$ is categorized (or directed) to one of the child nodes based on a simple statement (or condition) such as $x_j > v_k$ for certain feature $j$ and constant $v_k$. Since the entire decision-making process can be written as a decision rule with a set of propositional statements, decision trees are widely used in the industry for their transparency.

{\bf A Tree Ensemble} makes a prediction by combining decisions from multiple decision trees by averaging them. Formally, given $|T|$ decision trees $\{\mathcal{D}_1\cdots \mathcal{D}_{|T|}\}$ with tree contribution weight $\mathbf{w}\in {\rm I\!R}^{|T|}$, the prediction of the tree ensemble model is
\begin{equation}
   \underbrace{\hat{y} = \mathcal{M}(\mathbf{x}) = \sum_{t=1}^{|T|} w_{t} \mathcal{D}_{t}(\mathbf{x})}_{\textit{for regression}} \quad \textit{and} \quad \underbrace{\hat{y} =   \underset{c}{\arg\!\max} \mathcal{M}(\mathbf{x}) = \underset{c}{\arg\!\max} \sum_{t=1}^{|T|} w_{t} I[\mathcal{D}_{t}(\mathbf{x}) = c]}_{{\textit{for classification}}},
\end{equation}
where $I[\cdot]$ denotes the indicator function and $c \in C$ denotes the class index. This formulation applies to most of the well-known tree ensemble models, such as Random Forest~\cite{breiman2001random}, Boosted Trees~\cite{friedman2001greedy}, and XGBoost~\cite{chen2015xgboost}. In this paper, we consider adversarial robustness testing for tree ensemble models.

\section{Large-scale Adversarial Robustness Testing on Tree Ensembles}
% Problem Definition
% Our goal is to attack decision tree based classifiers, e.g. xgboost and random forest, with gradient based attacks. This requires estimating the gradient on the input of a decision tree wrt to its output. This document describes methods of 1) smoothing a decision tree; 2) calculating the full gradient of the smoothed tree; 3) efficiently sampling the gradient of the smoothed tree.
Given a predictive model $\hat{y} = \mathcal{M}(\mathbf{x})$, adversarial robustness testing aims to find an adversarial example $\mathbf{x}'$ (for each testing sample $\mathbf{x}$), which would cause the model to violate criteria tuple $(\Phi, \Psi, \epsilon, \delta)$ by allowing
\begin{equation}
\Psi(\mathcal{M}(\mathbf{x}'),\mathcal{M}(\mathbf{x})) > \delta \quad \textit{when} \quad \Phi(\mathbf{x}', \mathbf{x}) \leq \epsilon,
\end{equation}
where $\Phi$ denotes distance of inputs, $\Psi$ denotes distance of predictions, $\epsilon$ denotes perturbation criteria, and $\delta$ denotes the tolerance in prediction shift. In the literature, input distance $\Phi$ is usually the $l_{\infty}$ norm, whereas output distance $\Psi$ is the $l_1$ norm (absolute difference). Intuitively, adversarial robustness describes how well the model can preserve its prediction when the input data point is perturbed within a pre-defined perturbation bound. 

This paper proposes a large-scale adversarial robustness testing approach on tree ensemble models through iterative gradient descent. We first describe how to smooth the trees to support gradient descent. Then, we show how to conduct the adversarial example search on smoothed trees efficiently. 

\subsection{Tree Ensemble Smoothing through Branching Node Relaxation}
Among various white-box adversarial attack methods, gradient descent-based approaches usually show a significant advantage in effectiveness and computational efficiency~(with the existing toolboxes for auto-differentiation). To enable such an attack on tree ensemble models, we propose smoothing the tree ensemble to support auto-differentiation.

As reviewed in Section~\ref{sec:tree_ensemble}, the prediction of a tree ensemble is a linear combination of predictions from a set of decision trees. Since each decision tree can be factorized as a set of piece-wise branching nodes, we can smooth it by replacing the branching nodes with the softmax function~\footnote{While the smoothing operation supports multi-children branching, in practice, the internal nodes in a decision tree are usually binary branching nodes. In this paper, our description focuses on binary nodes.} Concretely, for an intermediate node $k\in \{1\cdots |K|\}$ of a tree $\mathcal{D}_{t}$ with two child nodes ($\mathcal{D}_{t,k}^{\textit{left}}$, $\mathcal{D}_{t,k}^{\textit{right}}$) and branching condition $x_j > v_k$ in the form of
\begin{equation}
    \mathcal{D}_{t,k}(\mathbf{x}) = 
    \begin{cases}
        \mathcal{D}_{t,k}^{\textit{left}}(\mathbf{x}) & x_j > v_k \\
        \mathcal{D}_{t,k}^{\textit{right}}(\mathbf{x}) & \textit{otherwise}\\
    \end{cases},
\end{equation}
we relax each intermediate node with a probabilistic distribution such that
\begin{equation}
   \tilde{\mathcal{D}}_{t,k}(\mathbf{x}) = \mathcal{Q}_{t,k}^{\textit{left}}(\mathbf{x})\tilde{\mathcal{D}}_{t,k}^{\textit{left}}(\mathbf{x}) + \mathcal{Q}_{t,k}^{\textit{right}}(\mathbf{x})\tilde{\mathcal{D}}_{t,k}^{\textit{left}}(\mathbf{x}),
\end{equation}
where the distribution can be expressed through a simple sigmoid function 
\begin{equation}
    \mathcal{Q}_{t,k}^{\textit{left}}(\mathbf{x}) = \textit{sigmoid}(\frac{x_j - v_k}{ \sigma_j}) \quad \textit{and} \quad \mathcal{Q}_{t,k}^{\textit{right}}(\mathbf{x}) = 1 - \mathcal{Q}_{t,k}^{\textit{left}}(\mathbf{x}).
\end{equation}
Here, we introduced the standard deviation $\sigma_j$ of feature $j$ in the training set into the sigmoid function to normalize the signals from all decision nodes.

With the above smoothing steps, the prediction of the smoothed tree ensemble $\tilde{\mathcal{M}}$ can be represented as
\begin{equation}
    \hat{y} = \underset{c}{\arg\!\max}\tilde{\mathcal{M}}(\mathbf{x}) = \underset{c}{\arg\!\max} \sum_{t=1}^{|T|} w_{t} \underbrace{\sum_{p=1}^{|P_t|}v_p\prod_{l=1}^{|L_p|}\mathcal{Q}_{t,l}^{p}(\mathbf{x})}_{\tilde{\mathcal{D}}_t(\mathbf{x})},
\label{eq:smoothed_prediction}
\end{equation}
where $P_t$ denotes the set of possible paths of a tree $t$, $l\in L_p$ denotes the $l$'th node in the decision path $p$, and $v_p\in C$ denotes the leaf node value of path $p$.

While our approach on smoothing a tree ensemble appears similar to the smoothing approach introduced in Soft Decision Tree~\cite{irsoy2012soft}, the fundamental difference is that we smooth the tree ensembles as a post-processing step to facilitate adversarial robustness testing instead of doing so to train a decision tree. 

\begin{figure}[t]
     \centering
     \begin{subfigure}[b]{0.25\linewidth}
         \centering
         \includegraphics[width=1.0\linewidth]{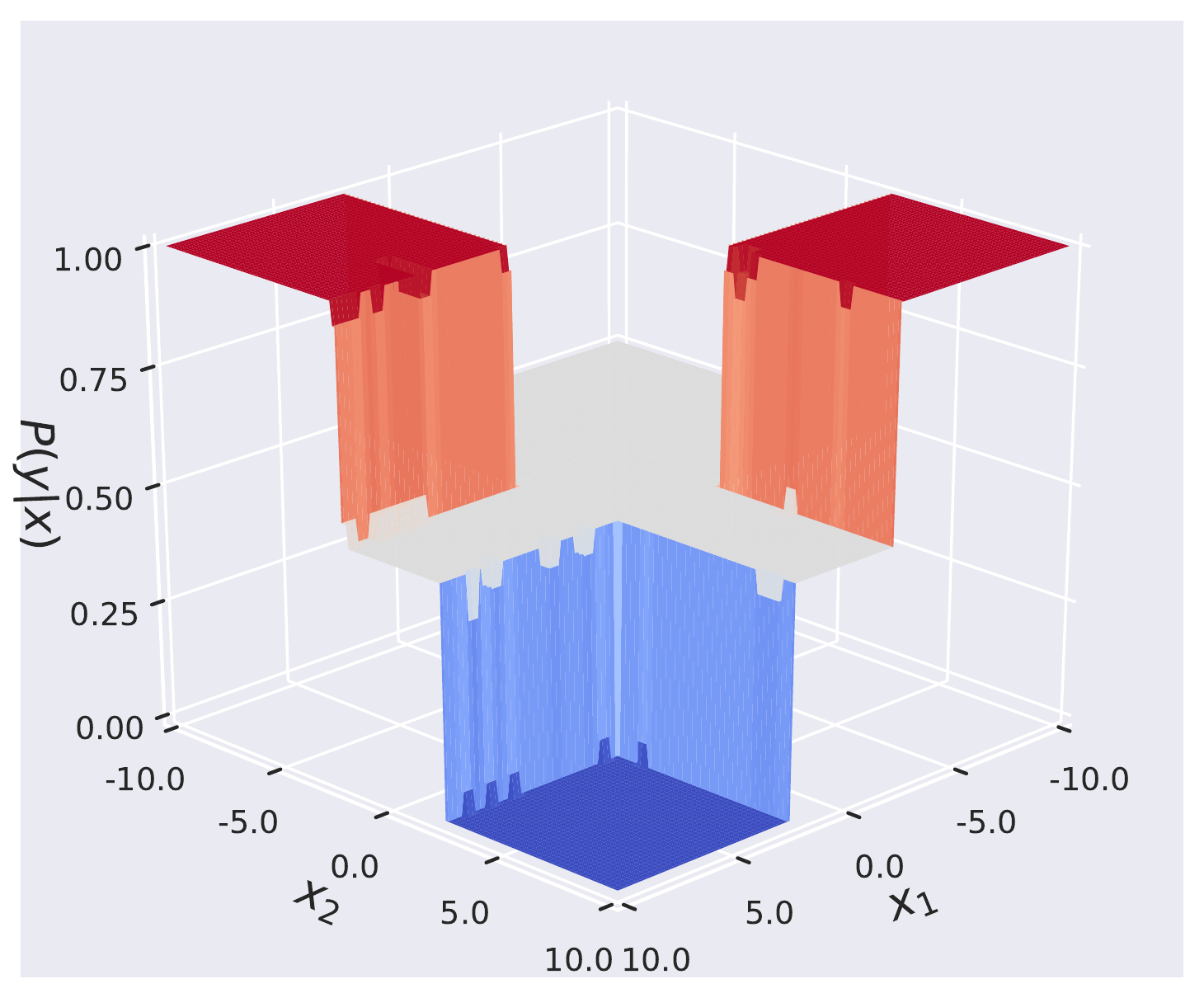}
         \caption{Original}
     \end{subfigure}
     \begin{subfigure}[b]{0.25\linewidth}
         \centering
         \includegraphics[width=1.0\linewidth]{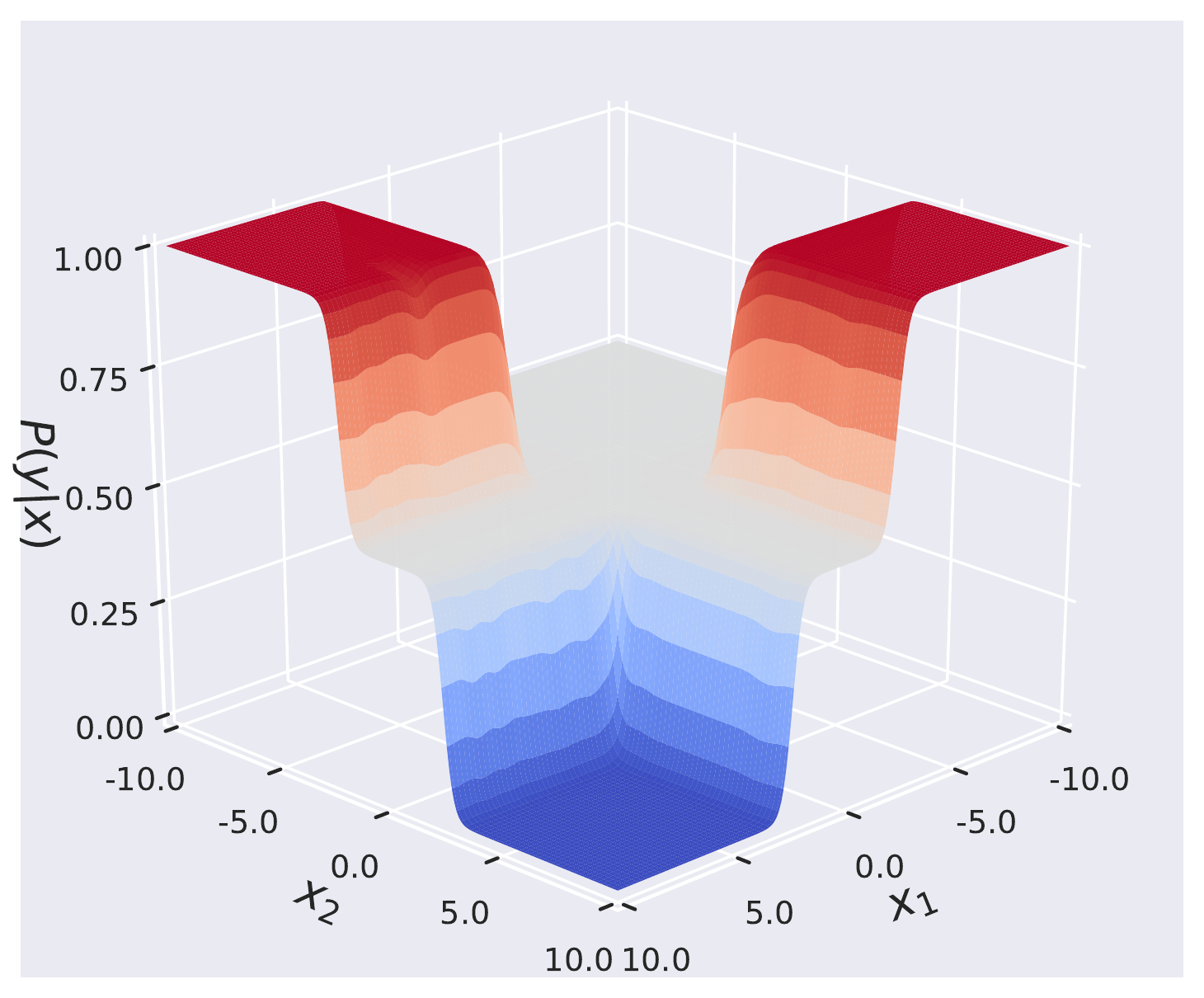}
         \caption{$\tau=0.01$}
     \end{subfigure}
          \begin{subfigure}[b]{0.25\linewidth}
         \centering
         \includegraphics[width=1.0\linewidth]{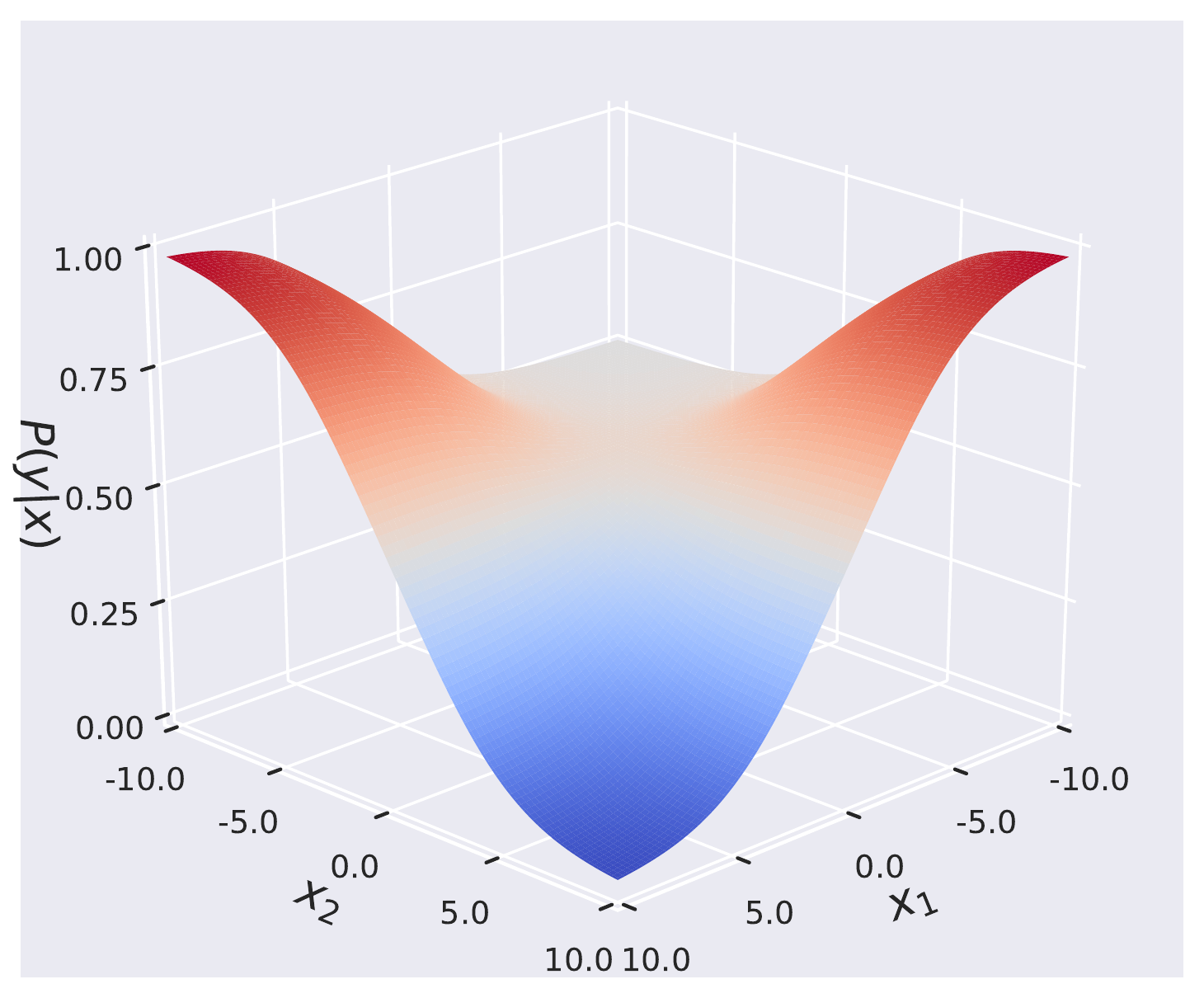}
         \caption{$\tau=0.1$}
     \end{subfigure}
    \caption{\textbf{Decision Surface Comparison between Tree Ensemble Smoothing and Original Model with Different Smoothing Temperature.} (a) The original tree ensemble. (b) low temperature approximation that maximizes the decision boundary preservation. (c) high temperature approximation with smoother decision surface that facilitate gradient descent.}
    \label{fig:tempered_sigmoid} 
\vspace{-5mm}
\end{figure}

\subsubsection{Balancing Gradient Descent Efficiency and Decision Surface Preservation}
% Why do we control the smoothed tree? -- Gradient Vanishing Problem
% How do we control? -- Temperature Hyper-parameter, default setting as a function of parameters? scheduler
While smoothing tree ensembles with sigmoid functions enables the usage of gradient descent, we note that the smoothed model may introduce two potential problems: 1) significant prediction gaps between the smoothed model and the original model, and 2) vanishing gradients due to the saturation of forward propagation over sigmoid functions.

To address the two potential problems, we introduce a hyper-parameter $\tau$ to control the activation of sigmoid functions (also called tempered sigmoids in the literature) such that
\begin{equation}
    \mathcal{Q}_{t,k}^{\textit{left}}(\mathbf{x}) = \textit{sigmoid}(\frac{x_j - v_k}{ \tau \sigma_j}). 
\end{equation}
Figure~\ref{fig:tempered_sigmoid} demonstrates the effect of controlling temperature hyper-parameter $\tau$. A higher temperature results in a smoother decision surface while severely increasing the approximation gap between the smoothed model and the original model. Conversely, a low temperature may not support adversarial example search since the surface exhibits piece-wise behaviour. In the limit when $\tau \rightarrow 0$, the smoothed tree ensemble will fall back to the original model.

\subsection{Gradient Descent based Adversarial Example Search on Smoothed Trees}
% General description of Adversarial Example Search
According to the Basic Iterative Method (BIM)~\cite{kurakin2016adversarial}, given data point $(\mathbf{x}, y)$, gradient descent based search aims to  create a potential adversarial example $\mathbf{x}'$ by iteratively maximizing the prediction cost function $\mathcal{J}(\mathcal{M},\mathbf{x},y)$ with
\begin{equation}
    \mathbf{x}^{(i+1)} = \mathbf{x}^{(i)} + \nabla_{\mathbf{x}} \mathcal{J}(\tilde{\mathcal{M}},\mathbf{x}^{(i)},y) \quad \textit{for} \quad i\in{1\cdots |I|},
\end{equation}
where $\mathbf{x}^{(0)} = \mathbf{x}$ as the starting point.
%As mentioned previously, adversarial robustness testing typically comes with the requirement that a valid adversarial example $\mathbf{x}'$ has to flip the model prediction with minimum confidence gap $\delta$ while keeping the input distance to original input $\mathbf{x}$ to be less than perturbation tolerance $\epsilon$. 
%\chris{I'm not sure you need the last sentence. You already said this earlier and it's a bit redundant.} 
% WARNING REDUNDANT!!
%In this paper, we halt the search based on two conditions: 1) the perturbation is over the tolerance $\Phi(\mathbf{x}^{(i)}, \mathbf{x}) >= \epsilon$. 2) the search is over the maximum number of iterations $i >= |I|$.

Using the gradient of Equation~\ref{eq:smoothed_prediction} directly in BIM requires traversing all the possible paths in the trees, which could be computationally intensive. Moreover, tree ensembles and their applications are different from neural networks: i) their decision surfaces are piece-wise constant, which reduces the efficiency of na\"ive gradient search; and ii) the inputs of the models may be not normalized and originate from very different distributions. Thus, using the same tolerance $\epsilon$ for inputs with different distributions could be misleading. In the subsequent sections, we propose some strategies to address these problems.

\begin{figure}[t]
     \centering
     \begin{subfigure}[b]{0.495\linewidth}
     \begin{subfigure}[b]{0.495\linewidth}
         \centering
         \includegraphics[width=1.0\linewidth]{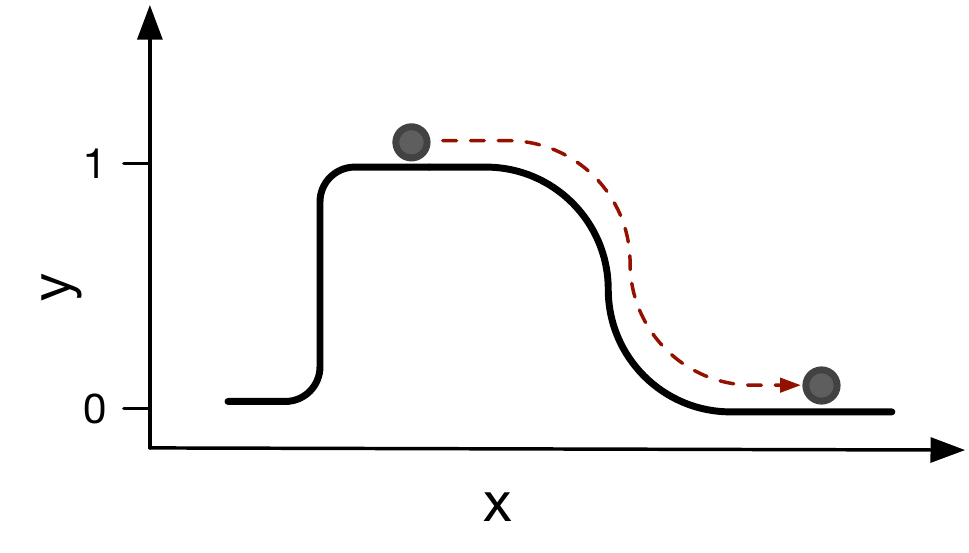}
         \caption{Pure Gradient}
     \end{subfigure}
     \begin{subfigure}[b]{0.495\linewidth}
         \centering
         \includegraphics[width=1.0\linewidth]{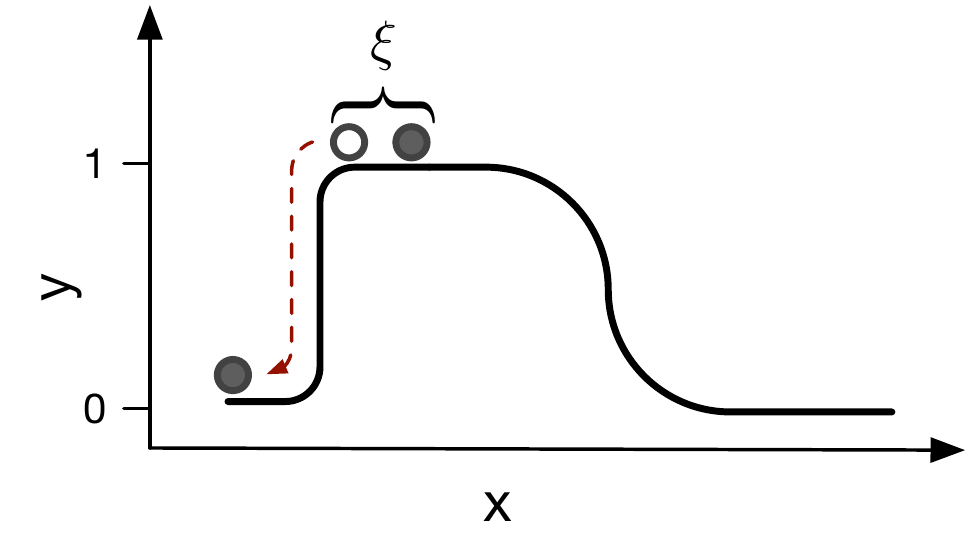}
         \caption{Noise Injected}
     \end{subfigure}
    \caption*{Conceptual Illustration}
    \end{subfigure}
    \begin{subfigure}[b]{0.495\linewidth}
         \begin{subfigure}[b]{0.495\linewidth}
         \centering
         \includegraphics[width=1.0\linewidth]{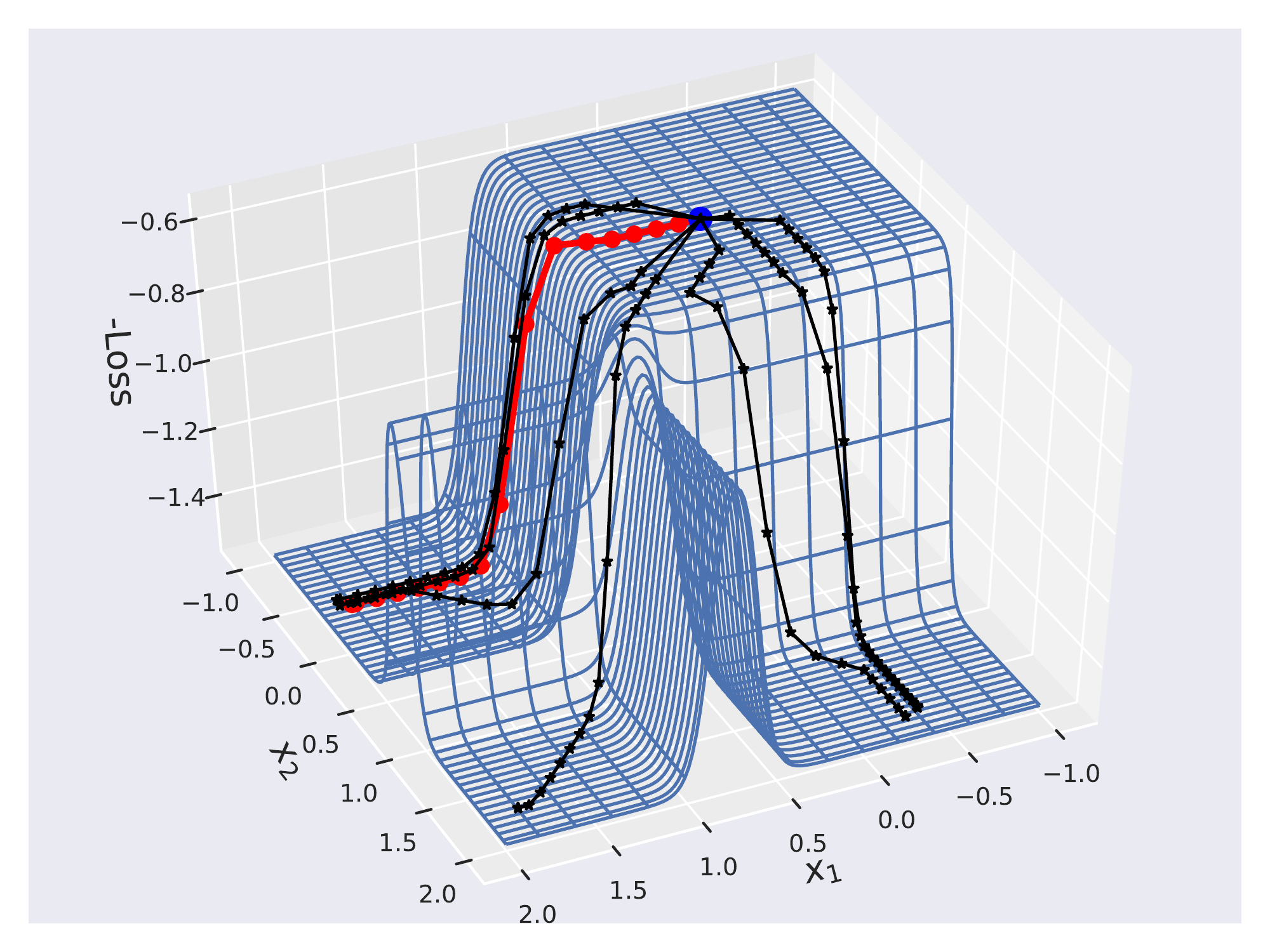}
         \caption{Sub-optimal}
     \end{subfigure}
     \begin{subfigure}[b]{0.495\linewidth}
         \centering
         \includegraphics[width=1.0\linewidth]{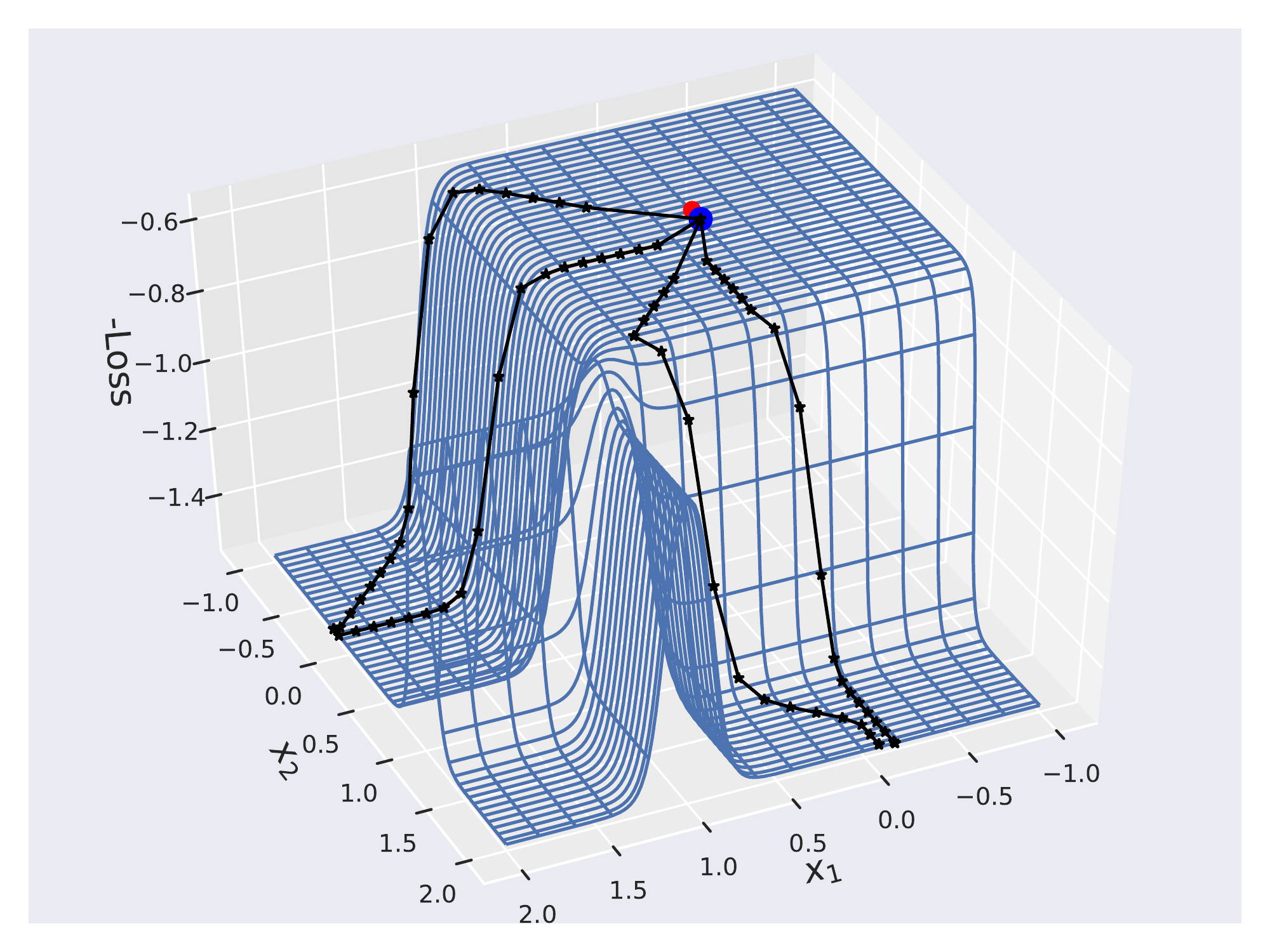}
         \caption{Stucked}
     \end{subfigure}
     \caption*{Actual Observation}
     \end{subfigure}
    \caption{\textbf{Adversarial Example Searching with/without Noise Injection on Smoothed Tree Ensemble.} a) Since smoothed tree ensembles preserve near piecewise decision surface, it may lead the search to sub-optimal direction. b) By injecting noise $\xi$ into the search, we may find adversarial examples with shorter traveling distance. c-d) Noise injection allows better coverage of searching (black trajectories), while pure gradient would either lead suboptimal solution or stuck on the platform (red trajectories).}
    \label{fig:gradient_direction}
\vspace{-5mm}
\end{figure}

\subsubsection{Noise Injected Gradient Descent and Search Coverage Maximization}
% Why do we need to inject noise during adversarial example search? -- Search space coverage
% Connect to the classic adversarial robustness testing literature.

The travel direction of the largest gradient is not necessarily the shortest path to finding an adversarial example, as shown in Figure~\ref{fig:gradient_direction}(a). Indeed, the shortest path to flip the decision could require climbing over a hill that is opposite to the gradient direction. While this problem does not hurt the effectiveness of adversarial testing on neural network models, we note it is a non-negligible issue when working with smoothed trees in our setting due to having a near piece-wise decision surface with many saddle points where the algorithm could get stuck. %\chris{can you expand a bit on why the piece wise nature can cause this problem or add anything else? This could be an interesting point to comment since it is not a problem that would occur in common AR literature with NNs.}

To mitigate this problem, we adopt Noise Injected Gradient Descent, where we add \textit{sparse} noise in each step of the optimization such that
\begin{equation}
    \mathbf{x}^{(i+1)} = \left[(1 \pm \bm{\xi}) \mathbf{x}^{(i)} + \nabla_{\mathbf{x}} \mathcal{J}(\tilde{\mathcal{M}},\mathbf{x}^{(i)},y)\right] \quad \textit{for} \quad i\in\{1\cdots |I|\},
\end{equation}
where $i$ denotes the iteration of the search, and we sample $\xi_j \sim \mathcal{N}(0, \lambda)$ for a sampled feature index $j$ to maximize the coverage of the adversarial example search. For other feature dimension $j'$, the value of $\xi_{j'}$ remains zero. Here, we introduce a hyper-parameter $\lambda$ that controls the noise level. Figure~\ref{fig:gradient_direction} (b) illustrates the purpose of injecting noise into the gradient. 

While many other whitebox attack algorithms also introduce noise during the adversarial example search~\cite{kurakin2016adversarial}, there are two differences between our approach and the others: 1) The noise introduced is NOT additive to the original value of $\mathbf{x}$ and thus keeps the scale of the feature value under consideration, which is particularly useful for applications where features are not of the same magnitude. 2) The noise is sparse and thus limits the noise injection to one dimension at a time. This helps improve the efficiency of noise injection since the movement to the closest adversarial example is always perpendicular to the decision boundary, which is near piece-wise in smoothed trees. 

% \begin{figure}[h]
% \centering
% \includegraphics[width=\textwidth]{figs/percentiles.png}
% \caption{\textbf{Normalizing features using quantile transforms:} the cumulative distribution function $\Gamma$ maps features uniformly into the range $[0,1]$, in which we define our perturbations.}
% \label{fig:percentile}
% \end{figure}

\subsubsection{Tree Sampling for Fast Adversarial Example Search}
\label{sec:sampling}
\vspace{-1mm}
% Why do we do fast approximation with tree sampling? -- Too slow to conduct the search.
% What is the gap between the fast approximation and the exhaustive search? -- Error analysis

% comparsion between subsection 2 and 3

%\wuga{Now, we describe an alternative whitebox attack method that is also based on the smoothed tree ensemble, which is more efficient than the exhaustive search approach discussed previously.} 

Calculating the gradient of a smoothed tree ensemble $\frac{\partial \tilde{\mathcal{M}}(\mathbf{x})}{\partial \mathbf{x}}$ based on the model description shown in Equation~\ref{eq:smoothed_prediction} is the computationally expensive. The operation requires traversing all the nodes in each of the trees $\mathcal{D}_t$ and leads to computational complexity $O(|K_t|+|L_t|)$ for a tree with $|K_t|$ nodes and depth $|L_t|$. In the worse case, we note $|K_t|+|L_t| \approx 2^{|L_t|}$. 

To mitigate the computational pressure, we propose sampling trajectories from each of the smoothed trees in a similar fashion to the original decision tree (only track one single path from all possible paths from root to leaves). Concretely, we aim to approximate the derivative of the smoothed tree ensemble $\nabla_{\mathbf{x}} \tilde{\mathcal{M}}(\mathbf{x})$ with a \textit{log-derivative} trick~\footnote{This log-derivative trick is commonly used in the reinforcement learning literature to avoid non-differentiable reward functions. E.g. policy gradient.} to enable sampling:
\begin{equation}
    \begin{aligned}
    \nabla_{\mathbf{x}} \tilde{\mathcal{M}}(\mathbf{x}) & = \nabla_{\mathbf{x}} \sum_{t=1}^{|T|} w_{t} \sum_{p=1}^{|P_t|}v_p\prod_{l=1}^{|L_p|}\mathcal{Q}_{t,l}^{p}(\mathbf{x})  = \sum_{t=1}^{|T|} w_{t} \sum_{p=1}^{|P_t|}v_p\prod_{l=1}^{|L_p|} \mathcal{Q}_{t,l}^{p}(\mathbf{x}) \nabla_{\mathbf{x}} \log \mathcal{Q}_{t,l}^{p}(\mathbf{x})\\
    & \approx \sum_{t=1}^{|T|} w_{t}  \underbrace{\left(\sum_{l=1}^{|L_p|}\nabla_{\mathbf{x}} \log \mathcal{Q}_{t,l}^{p}(\mathbf{x}) v_p\right)}_{\textit{gradient of sampled path}} \quad \textit{for} \quad p\sim \textit{Mult}(P_t, \mathcal{Q}_t),\\
    \end{aligned}
\end{equation}
% \underbrace{\left(\prod_{l=1}^{|L_p|} \mathcal{Q}_{t,l}^{p}(\mathbf{x}) \right)}_{\textit{probability of sampled path}}
which allows us to obtain an unbiased estimate of the derivative $\nabla_{\mathbf{x}} \tilde{\mathcal{M}}(\mathbf{x})$ by sampling single path $p$ for each tree from a multinomial distribution $\mathcal{Q}_t = \{\prod_{l=1}^{|L_p|} \mathcal{Q}_{t,l}^{1}(\mathbf{x}), \cdots,\prod_{l=1}^{|L_p|} \mathcal{Q}_{t,l}^{|P_t|}(\mathbf{x})\}$.

\begin{figure}[t]
     \centering
     \begin{subfigure}[b]{0.495\linewidth}
     \begin{subfigure}[b]{0.495\linewidth}
         \centering
         \includegraphics[width=1.0\linewidth]{figs/tree_exhaustive_tmp_0.01.pdf}
         \caption{Exhaustive}
     \end{subfigure}
     \begin{subfigure}[b]{0.495\linewidth}
         \centering
         \includegraphics[width=1.0\linewidth]{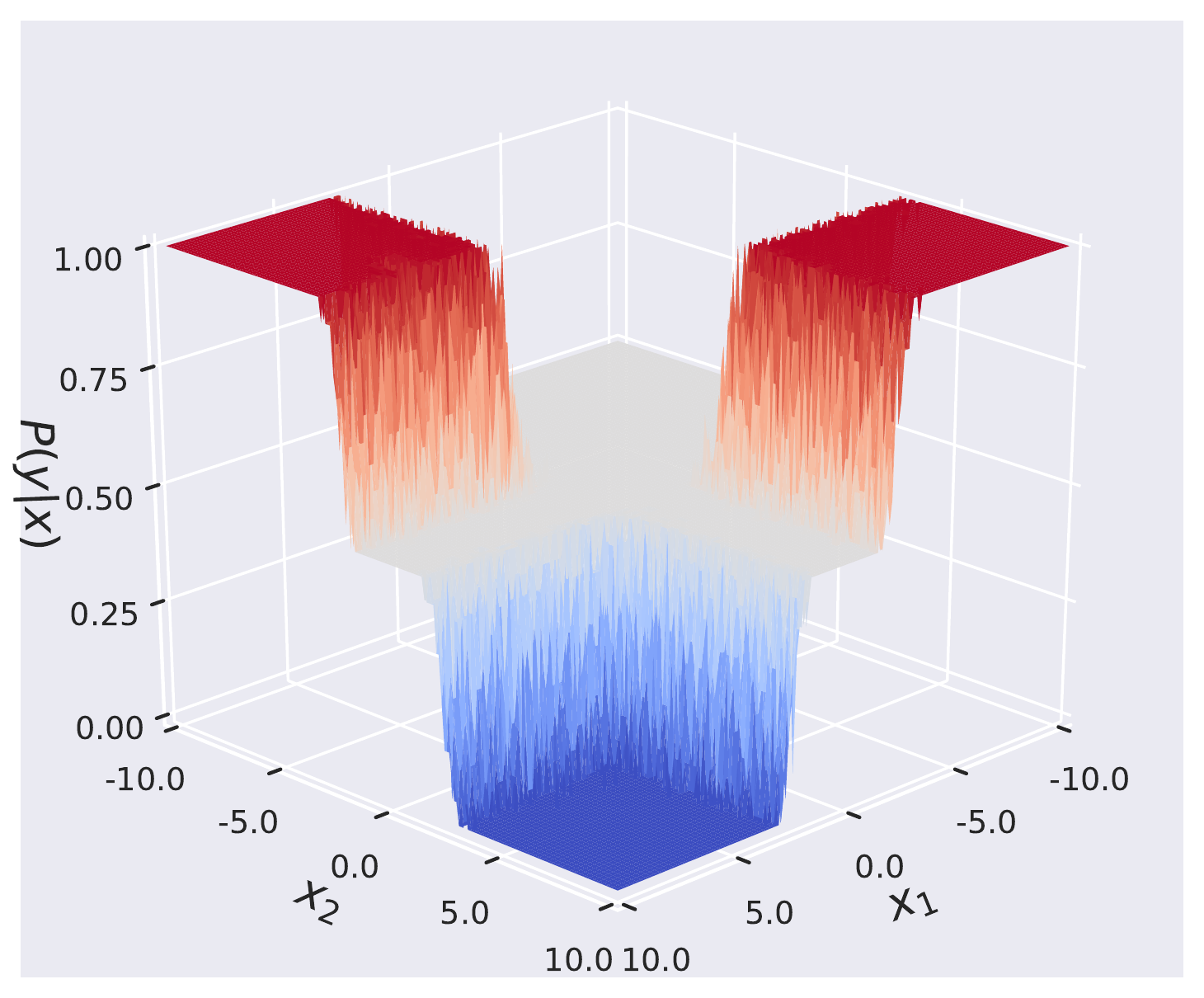}
         \caption{Sampled}
     \end{subfigure}
    \caption*{$\tau = 0.01$}
    \end{subfigure}
    \begin{subfigure}[b]{0.495\linewidth}
         \begin{subfigure}[b]{0.495\linewidth}
         \centering
         \includegraphics[width=1.0\linewidth]{figs/tree_exhaustive_tmp_0.1.pdf}
         \caption{Exhaustive}
     \end{subfigure}
     \begin{subfigure}[b]{0.495\linewidth}
         \centering
         \includegraphics[width=1.0\linewidth]{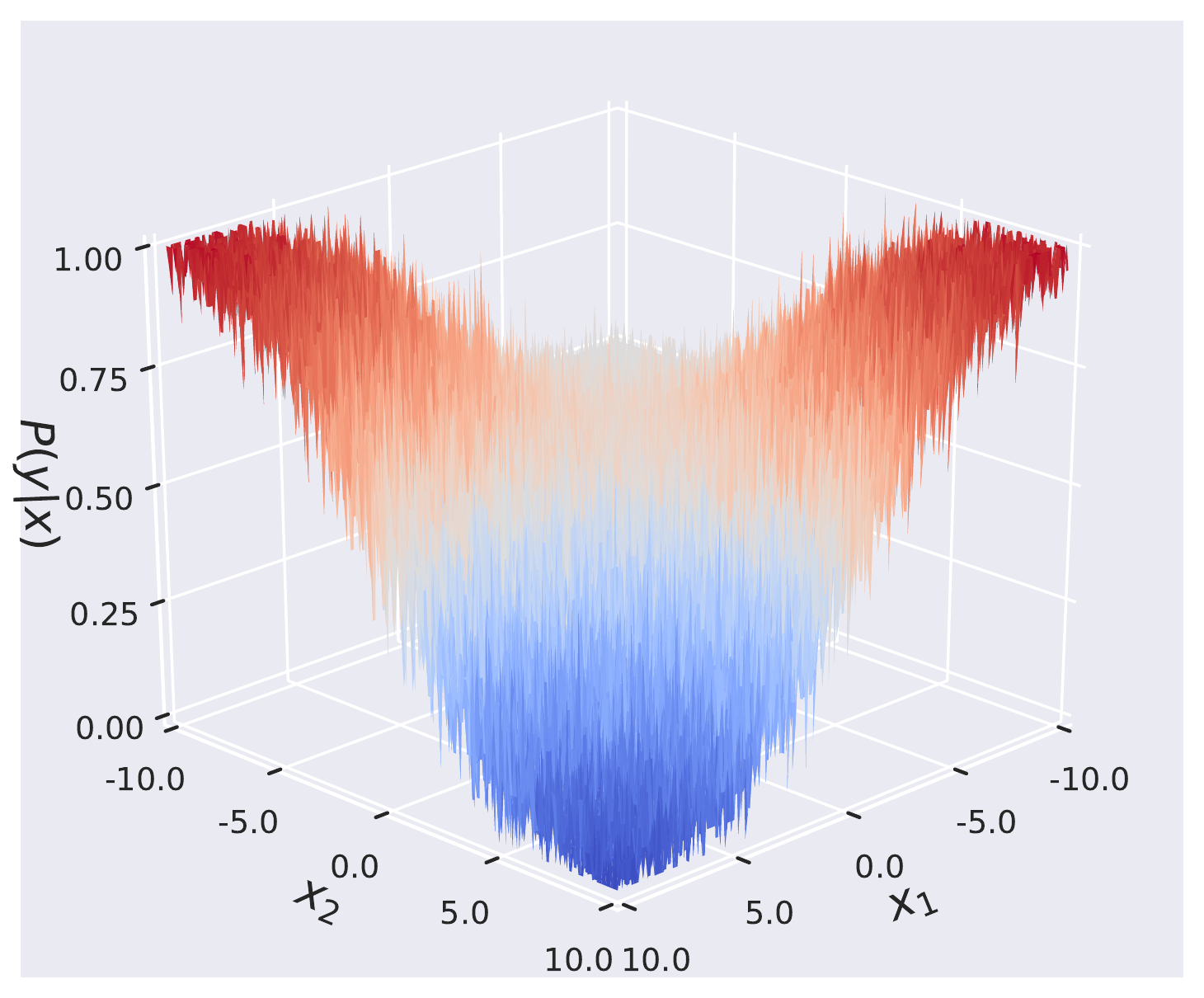}
         \caption{Sampled}
     \end{subfigure}
     \caption*{$\tau = 0.1$}
     \end{subfigure}
    \caption{\textbf{Decision Surface Comparison between Exhaustive Search and Sampled Search with Different Smoothing Temperature.} (a-b) low temperature approximation that maximizes the decision boundary preservation. (c-d) high temperature approximation with smoother decision surface that facilitate gradient descent.}
    \label{fig:exhaustive_vs_sampled}
\vspace{-5mm}
\end{figure}

To reduce the variance of the estimate, we can also choose to sample multiple times and take the numerical expectation. However, from our experiments, we empirically show that sampling once is sufficient to achieve a reasonable approximation, as shown in Figure~\ref{fig:exhaustive_vs_sampled}. Our conjecture of the observation is that we ensemble many trees in the tree ensemble model which smooths the noise introduced by the sampling.

\subsubsection{Feature Dependent Perturbation Range}
% Why $L-\infity$ along is not good idea for tree ensemble? 
%-- 1) work on tabular data (with different feature magnitude) -- discrete!!
%-- 2) feature importance are dramatically different with each other.
%-- 3) normalization / quantile transformer

Many existing works use a universal $l_\infty$ norm to define the distance between the original input and adversarial example up to tolerance $\epsilon$. While using a universal distance metric and tolerance is a reasonable setup in an experimental environment, we note that the synthetic setting would result in a misleading conclusion for predictive tasks whose input are tabular data, where we expect the features of the data to have different ranges, distributions, and interpretations. This is particularly critical task to address for tree based models, since the training data for tree ensemble models is not necessarily normalized (which is different from deep learning models). 

In this paper, we propose allowing the input perturbation range to be automatically adjusted for each feature based on feature statistics. 

Assuming there is an unkonwn Cumulative Density Function~(CDF) $\mathcal{F}_j$ for each feature $j$, where any observed feature assignment $x_j$ is a sample from the CDF such that
\begin{equation}
x_j = \mathcal{F}_j^{-1}(q) \quad \textit{and} \quad q\sim \mathcal{U}(0,1).
\end{equation}
Here, $\mathcal{U}$ denotes a uniform distribution. We propose bounding the feature perturbation in the range 
\begin{equation}
    x_j^{(i+1)} \in [\mathcal{F}^{-1}(\mathcal{F}(x_j)^{(i)}-\epsilon), \mathcal{F}^{-1}(\mathcal{F}(x_j)+\epsilon) ]
\end{equation}
with hyper-parameter $\epsilon \in \mathcal{U}(0,1)$. Intuitively, the above operation enables a uniform perturbation (controlled by $\epsilon$) with implicit feature normalization.

As the CDF is inaccessible in practice, we approximate the CDF empirically (ECDF) by 1) sorting feature observations in the training data, 2) sampling data with their percentile scores (i.e. the indices of the sorted observation list), and 3) linearly interpolating these feature values and percentile scores.

\vspace{-1mm}
\section{Experiment and Evaluation}
\vspace{-1mm}
% Experiment question list:
% Performance
% Timing/Memory
% High temperature vs Low temperature
% Noise vs No-noise
Now we proceed to evaluate the proposed adversarial robustness testing approach to answer the following questions:
\begin{itemize}[leftmargin=10pt, itemsep=0em, topsep=0pt]
    \item {\bf RQ1:} How effective is the proposed approach in terms of searching for adversarial examples on tree ensembles compared to existing state-of-the-art approaches (e.g. blackbox attacks) and random perturbations?
    \item {\bf RQ2:} How efficient is the proposed approach in terms of computational cost and time?
    \item {\bf RQ3:} What is the effect of tuning the temperature hyper-parameter during model smoothing?
    \item {\bf RQ4:} How does noise injection impact the search results?
    \item {\bf RQ5:} Does the trajectory sampling approach work well compared to the exhaustive search?
    %\item {\bf RQ6:} How well do the proposed approaches compare to blackbox adversarial attacks approaches? \chris{you mention blackbox attack in RQ1 also, is this different for this RQ?}
\end{itemize}

\begin{table}[t!]
\caption{{\bf Performance Comparison of Adversarial Robustness Testing.} Here we measure the prediction accuracy after adversarial attack. Lower is better. Results are collected from 3-fold cross-validation, and error shows standard derivation.}
%\rowcolors{2}{gray!10}{white}
\resizebox{\textwidth}{!}{%
\begin{tabular}{c|c|c|c|c|c|c|c|c}
\toprule
Ensemble& Data Name& Original & Tolerance & STA-Exhaustive & STA-Sampling & GenAttack & NES & Random\\
\midrule
\multirow{9}{*}{Forest}&\multirow{3}{*}{German Credit}& \multirow{3}{*}{$71.89\% \pm 1.83\%$}& $\epsilon=0.2$ & \textcolor[rgb]{0.141,0.153,0.161}{$44.80 \pm 2.72$} & $45.60 \pm 2.73$ & $42.90 \pm 3.62$ & $45.60\pm 2.82$ & $70.54 \pm 1.59$ \\
&&&$\epsilon=0.5$ & $25.79 \pm 12.22$ & $26.39 \pm 12.67$ & $22.69 \pm 13.26$ & $26.59 \pm 12.27$ & $66.89 \pm 0.86$ \\
&&&$\epsilon=0.8$ & $22.49 \pm 11.59$ & $22.39 \pm 11.33$ & $17.69 \pm 13.64$ & $23.39 \pm 11.83$ & $64.39 \pm 2.64$ \\
\cmidrule(lr){2-9}
&\multirow{3}{*}{Adult Salary}& \multirow{3}{*}{$77.59\% \pm 0.27\%$}&$\epsilon=0.2$ & \textcolor[rgb]{0.141,0.153,0.161}{$7.07 \pm 0.16$} & $7.11 \pm 0.17$ & $7.63 \pm 0.22$ & $7.24 \pm 0.17$ & $43.71 \pm 1.98$ \\
&&&$\epsilon=0.5$ & $0.54 \pm 0.07$ & $0.55 \pm 0.05$ & $0.58 \pm 0.02$ & $0.76 \pm 0.02$ & $32.81 \pm 1.82$ \\
&&&$\epsilon=0.8$ & $0.18 \pm 0.12$ & $0.19 \pm 0.11$ & $0.14 \pm 0.13$ & $0.43 \pm 0.13$ & $29.89 \pm 1.87$ \\
\cmidrule(lr){2-9}
% \multirow{3}{*}{Digit}& \multirow{3}{*}{$88.03\% \pm 0.61\%$}&$\epsilon=0.2$ & \textcolor[rgb]{0.141,0.153,0.161}{$15.47\pm 1.91$} & $17.92 \pm 1.75$ & $13.68 \pm 1.03$ & $17.25 \pm 1.88$ & $74.04 \pm 0.73$ \\
% &&$\epsilon=0.5$ & $0 \pm 0$ & $0 \pm 0$ & $0 \pm 0$ & $0 \pm 0$ & $44.60 \pm 2.32$ \\
% &&$\epsilon=0.8$ & $0 \pm 0$ & $0 \pm 0$ & $0 \pm 0$ & $0 \pm 0$ & $17.62 \pm 0.09$ \\
% \midrule
&\multirow{3}{*}{Breast Cancer}& \multirow{3}{*}{$95.43\% \pm 1.51\%$}&$\epsilon=0.2$ & \textcolor[rgb]{0.141,0.153,0.161}{$7.38 \pm 2.29$} & $8.97 \pm 3.01$ & $8.09 \pm 2.89$ & $10.55 \pm 2.42$ & $89.91 \pm 13.99$ \\
&&&$\epsilon=0.5$ & $0.17 \pm 0.25$ & $0.17 \pm 0.25$ & $ 0.17 \pm 0.25$ & $0.17 \pm 0.25$ & $51.88 \pm 0.25$ \\
&&&$\epsilon=0.8$ & $0 \pm 0$ & $0 \pm 0$ & $0 \pm 0$ & $0 \pm 0$ & $39.69 \pm 0.55$ \\
\midrule
\multirow{9}{*}{XGBoost}&\multirow{3}{*}{German Credit}& \multirow{3}{*}{$75.29\% \pm 3.13\%$}& $\epsilon=0.2$ & \textcolor[rgb]{0.141,0.153,0.161}{$42.29 \pm 8.56$} & $42.29 \pm 8.53$ & $25.39 \pm 7.32$ & $41.49\pm 8.80$ & $72.91 \pm 2.24$ \\
&&&$\epsilon=0.5$ & $18.19 \pm 11.83$ & $17.79 \pm 11.52$ & $14.29 \pm 9.25$ & $22.39 \pm 12.01$ & $67.19 \pm 5.15$ \\
&&&$\epsilon=0.8$ & $14.39 \pm 9.33$ & $14.59 \pm 9.27$ & $8.49 \pm 5.81$ & $18.89 \pm 11.53$ & $61.64 \pm 8.43$ \\
\cmidrule(lr){2-9}
&\multirow{3}{*}{Adult Salary}& \multirow{3}{*}{$85.28\% \pm 1.68\%$}&$\epsilon=0.2$ & \textcolor[rgb]{0.141,0.153,0.161}{$0.76 \pm 0.02$} & $0.87 \pm 0.03$ & $0.82 \pm 0.02$ & $0.98 \pm 0.02$ & $44.66 \pm 0.45$ \\
&&&$\epsilon=0.5$ & $0.42 \pm 0.02$ & $0.42 \pm 0.05$ & $0.13 \pm 0.04$ & $0.66 \pm 0.01$ & $32.68 \pm 0.02$ \\
&&&$\epsilon=0.8$ & $0.42 \pm 0.0$ & $0.41 \pm 0.0$ & $0.05 \pm 0.03$ & $0.63 \pm 0.09$ & $28.78 \pm 0.19$ \\
\cmidrule(lr){2-9}
% \multirow{3}{*}{Digit}& \multirow{3}{*}{$91.76\% \pm 2.45\%$}&$\epsilon=0.2$ & \textcolor[rgb]{0.141,0.153,0.161}{$5.98 \pm 1.85$} & $5.67 \pm 2.81$ & $5.17 \pm 2.75$ & $12.57 \pm 4.12$ & $77.51 \pm 0.10$ \\
% &&$\epsilon=0.5$ & $0 \pm 0$ & $0 \pm 0$ & $0 \pm 0$ & $0 \pm 0$ & $40.95 \pm 2.26$  \\
% &&$\epsilon=0.8$ & $0 \pm 0$ & $0 \pm 0$ & $0 \pm 0$ & $0 \pm 0$ & $16.74 \pm 0.08$ \\
% \midrule
&\multirow{3}{*}{Breast Cancer}& \multirow{3}{*}{$95.95\% \pm 0.89\%$}&$\epsilon=0.2$ & \textcolor[rgb]{0.141,0.153,0.161}{$7.56 \pm 2.23$} & $7.73 \pm 2.18$ & $7.38 \pm 2.41$ & $9.14 \pm 2.39$ & $84.53 \pm 2.51$ \\
&&&$\epsilon=0.5$ & $0.17 \pm 0.25$ & $0.17 \pm 0.25$ & $0 \pm 0.17 \pm 0.25$ & $0.17 \pm 0.25$ & $49.81 \pm 2.53$ \\
&&&$\epsilon=0.8$ & $0 \pm 0$ & $0 \pm 0$ & $0 \pm 0$ & $0 \pm 0$ & $39.78 \pm 1.01$\\
\bottomrule
\end{tabular}}
\label{table:adverarial_attack_performance}
\vspace{-5mm}
\end{table}
\subsection{Experimental Settings}
\label{sec:experimental_settings}
We evaluate the proposed adversarial robustness testing approach on multiple pre-trained Random Forest and XGboost models learned from multiple UCI datasets. All of the random forest models have 100 estimators with maximum tree depth 4. Other hyper-parameters of the tree ensembles remain default values. To evaluate performance, we show the accuracy degradation as evidence of the effectiveness of the adversarial attack. To evaluate the inference efficiency, we show the running time as the metric of comparison. 

In the experiments, we call our proposed approaches STA-Exhaustive and STA-Sampling. Here, STA stands for Smoothed Tree Attack. STA-Exhaustive denotes a whitebox attack that exhaustively searches adversarial examples by collecting gradients from all decision trajectories of smoothed trees. In contrast, STA-Sampling denotes the sampling-based whitebox attack we described in Section~\ref{sec:sampling} that reduces the inference time by sampling a single trajectory for each tree. The other candidate approaches in the experiments are GenAttack~\cite{alzantot2019genattack}, NES~\cite{ilyas2018black}, and Random, where Random denotes uniformly random perturbation on inputs to serves as a baseline attacking approach. 
%\chris{you may need to explain a bit what you by random perturbation, are you sampling according to a distribution? uniform?}

\subsection{Performance Evaluation on Benchmark Datasets}
% target models: Random Forest, XGBoost, LGBM
% baselines:
% random perturbation
% combinational optimization
% blackbox attack
% decision tree specific attacks -- do we have implementations for this? we have click paper
% cite MILP results directly on table. [Use * indicate we do direct refer] 
We evaluate the effectiveness of the proposed methods (STA-Exhaustive and STA-Sampling) in terms of generating adversarial examples given a perturbation tolerance. Table~\ref{table:adverarial_attack_performance} shows the experiment results. Here, we highlight our observations:
\begin{itemize}[leftmargin=10pt, itemsep=0pt, partopsep=0pt, topsep=0pt]
    \item The proposed approaches are effective in terms of looking for adversarial examples. When comparing to the random search baseline, there is a significant performance gap between STAs and Random.
    \item The proposed approaches show performance competitive with the state-of-the-art approaches, GenAttack and NES. In multiple cases (e.g. XGBoost trained on Adult Salary), the proposed approaches show significantly better performance than NES. Here, we want to highlight that the computational cost of GenAttack is exponentially more expensive than our proposed approaches, as we will show next.
    \item The sampling-based approximation shows a slightly worse performance than the exhaustive approach. However, such performance degradation is not statistically significant as their confidence intervals heavily overlap with each other.
    \item The proposed approaches show stable performance on both Random Forest and XGBoost models.
\end{itemize}

\subsection{Computational Efficiency against Combinational Optimization}
\label{sec:exp_computational_efficiency}
\begin{figure}[t]
     \centering
     \begin{subfigure}[b]{\linewidth}
     \begin{subfigure}[b]{0.32\linewidth}
         \centering
         \includegraphics[width=1.0\linewidth]{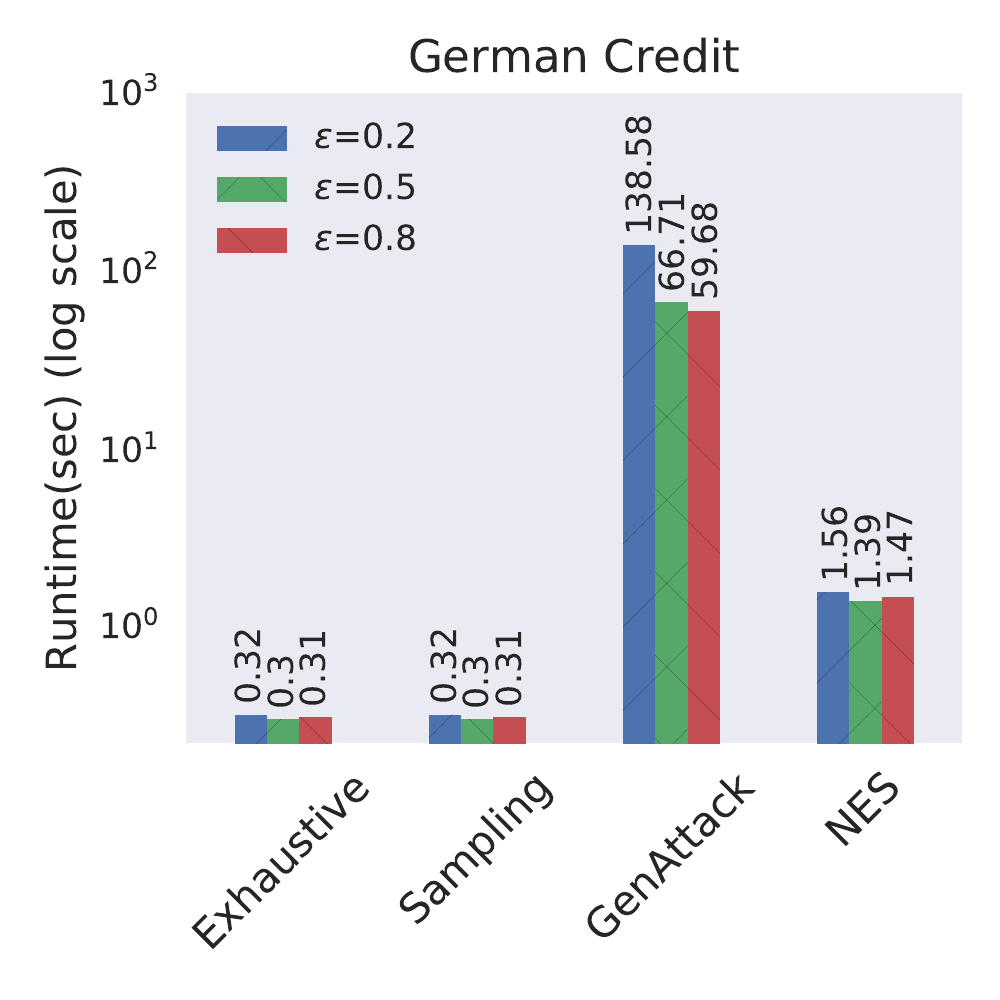}
     \end{subfigure}
     \begin{subfigure}[b]{0.32\linewidth}
         \centering
         \includegraphics[width=1.0\linewidth]{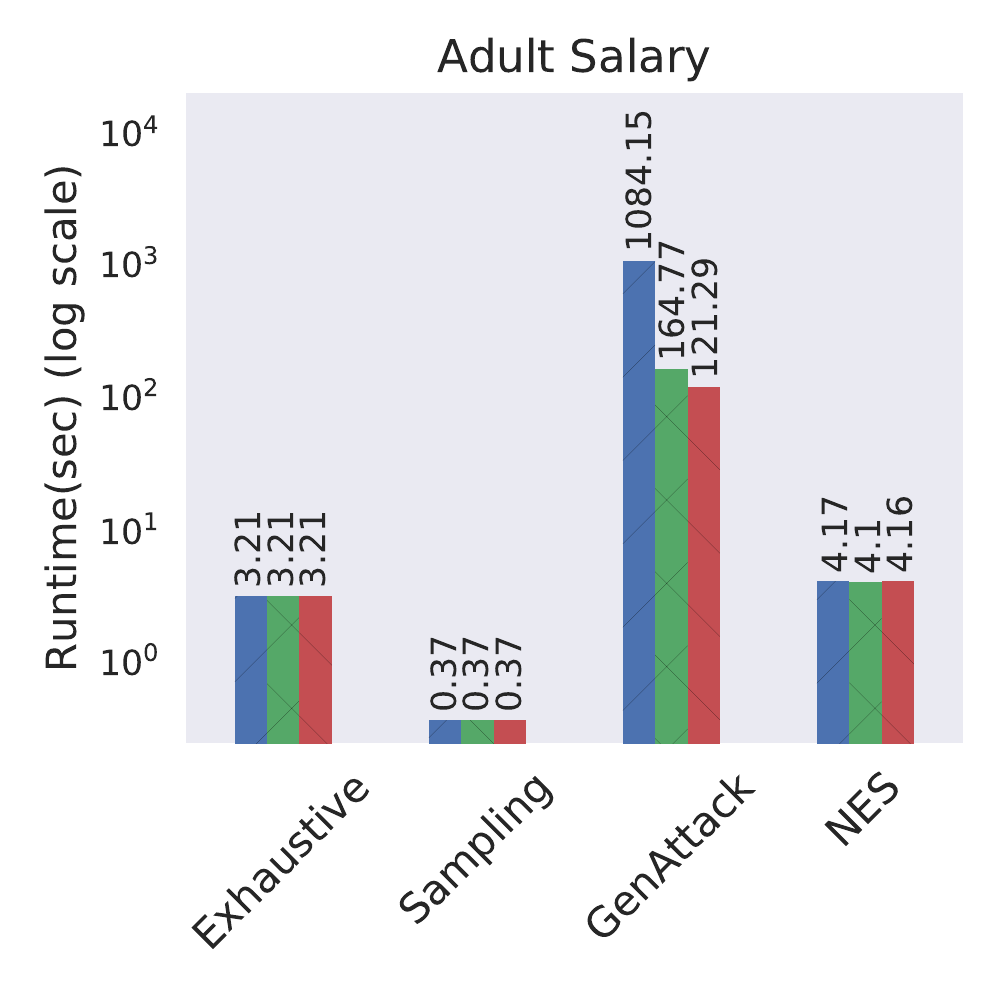}
     \end{subfigure}
    %      \begin{subfigure}[b]{0.32\linewidth}
    %      \centering
    %      \includegraphics[width=1.0\linewidth]{figs/rf_digits_time.eps}
    %      \caption{Exhaustive}
    %  \end{subfigure}
     \begin{subfigure}[b]{0.32\linewidth}
         \centering
         \includegraphics[width=1.0\linewidth]{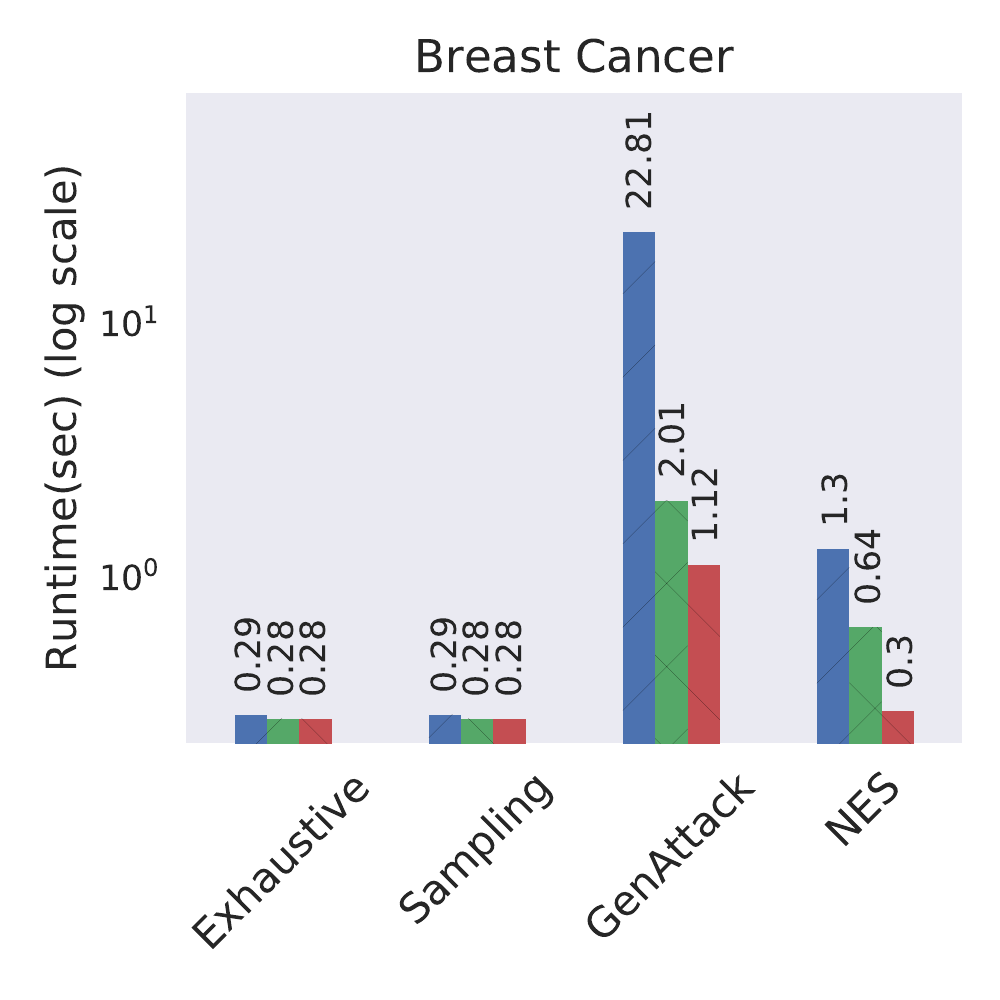}
     \end{subfigure}
     \vspace{-3mm}
     \caption{Random Forest}
     \end{subfigure}
     \begin{subfigure}[b]{\linewidth}
     \begin{subfigure}[b]{0.32\linewidth}
         \centering
         \includegraphics[width=1.0\linewidth]{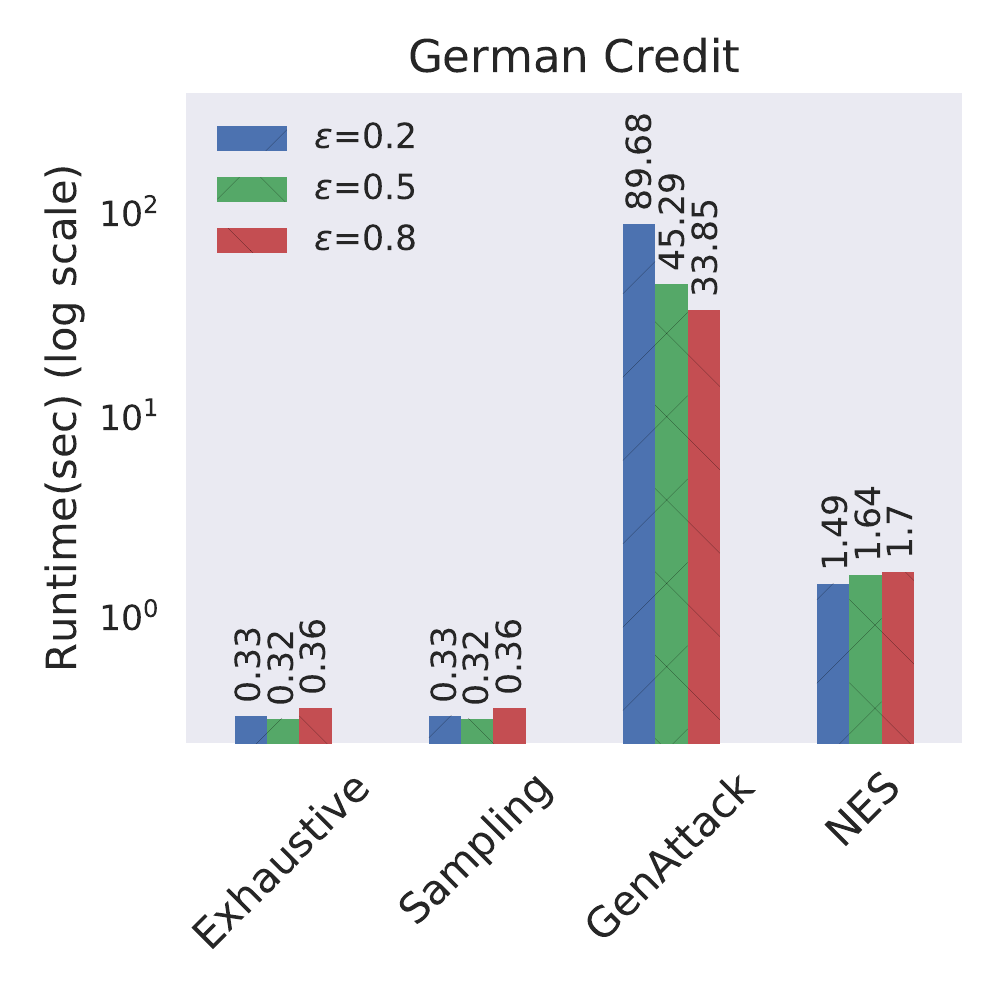}
     \end{subfigure}
     \begin{subfigure}[b]{0.32\linewidth}
         \centering
         \includegraphics[width=1.0\linewidth]{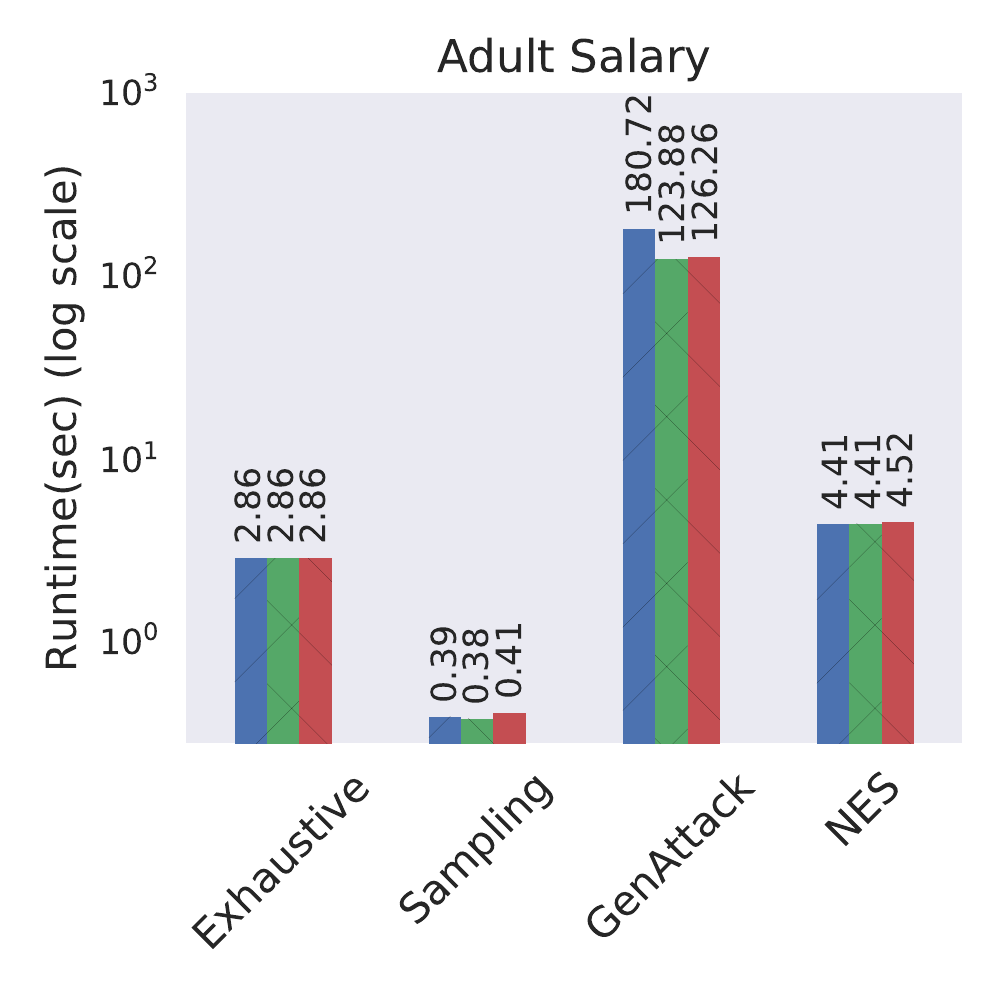}
     \end{subfigure}
    %      \begin{subfigure}[b]{0.32\linewidth}
    %      \centering
    %      \includegraphics[width=1.0\linewidth]{figs/xgb_digits_time.eps}
    %      \caption{Exhaustive}
    %  \end{subfigure}
     \begin{subfigure}[b]{0.32\linewidth}
         \centering
         \includegraphics[width=1.0\linewidth]{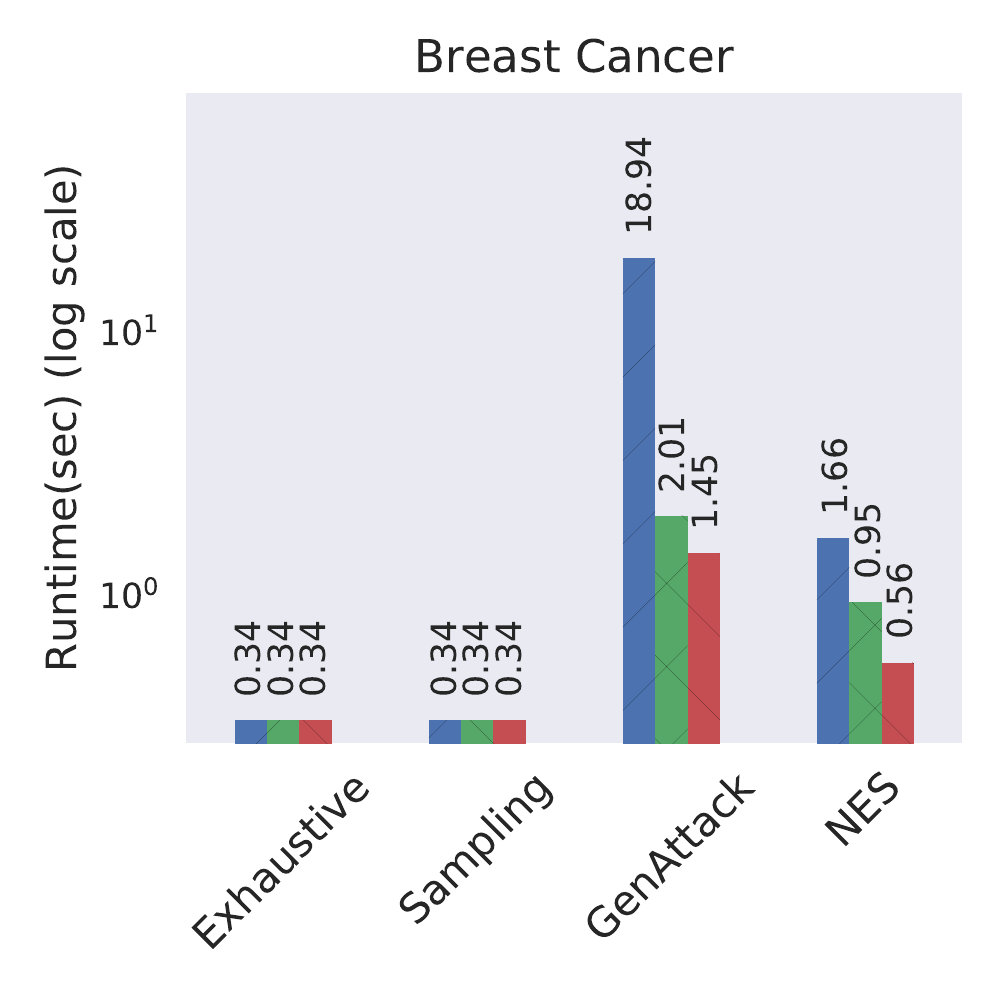}
     \end{subfigure}
     \vspace{-3mm}
     \caption{XGBoost}
     \end{subfigure}
    \caption{\textbf{Search time comparison between proposed and existing approaches.} We use a log scale for the comparison as the GenAttack shows exponentially more time consumption than other approaches. For bars, lower is better. Legends~(colours) show different perturbation tolerance. Values on top of bars are the actual run time in seconds. Lower is better. We omit error bars as the variance is negligible comparing to the significant run time gap among candidates.}
    \label{fig:run_time_comparison}
    \vspace{-5mm}
\end{figure}

\vspace{-2mm}
In this experiment, we show the computational efficiency of the proposed approaches. Figure~\ref{fig:run_time_comparison} shows the run time estimation that was conducted on Random Forest models. Experiments are done on a cluster with NVDIA Titan V GPUs.
Here, we list two important observations:
\begin{itemize}[leftmargin=10pt, itemsep=0em, topsep=0pt]
    \item the exhaustive variant is more than ten times faster than the NES approach and over 100 times faster than the GenAttack with similar performance. The sampling-based approach is even faster than the exhaustive variant (2-8 times faster depending on the application domain.). 
    \item For GenAttack, a smaller perturbation tolerance $\epsilon$ usually results in longer run time as it is hard to trigger early stopping due to the difficulty of searching for adversarial example with a small tolerance. In contrast, the efficiency of the proposed approaches is not sensitive to the size of epsilon.
    \item Combining the observations of Figure~\ref{fig:run_time_comparison} and Table~\ref{table:adverarial_attack_performance}, we note that the proposed approaches show a significant advantage in terms of efficiency while maintaining competitive adversarial attack performance, which make them more practical for large-scale testing purposes.
\end{itemize}

\begin{figure}[t]
     \centering
     \begin{subfigure}[b]{0.32\linewidth}
         \centering
         \includegraphics[width=0.9\linewidth]{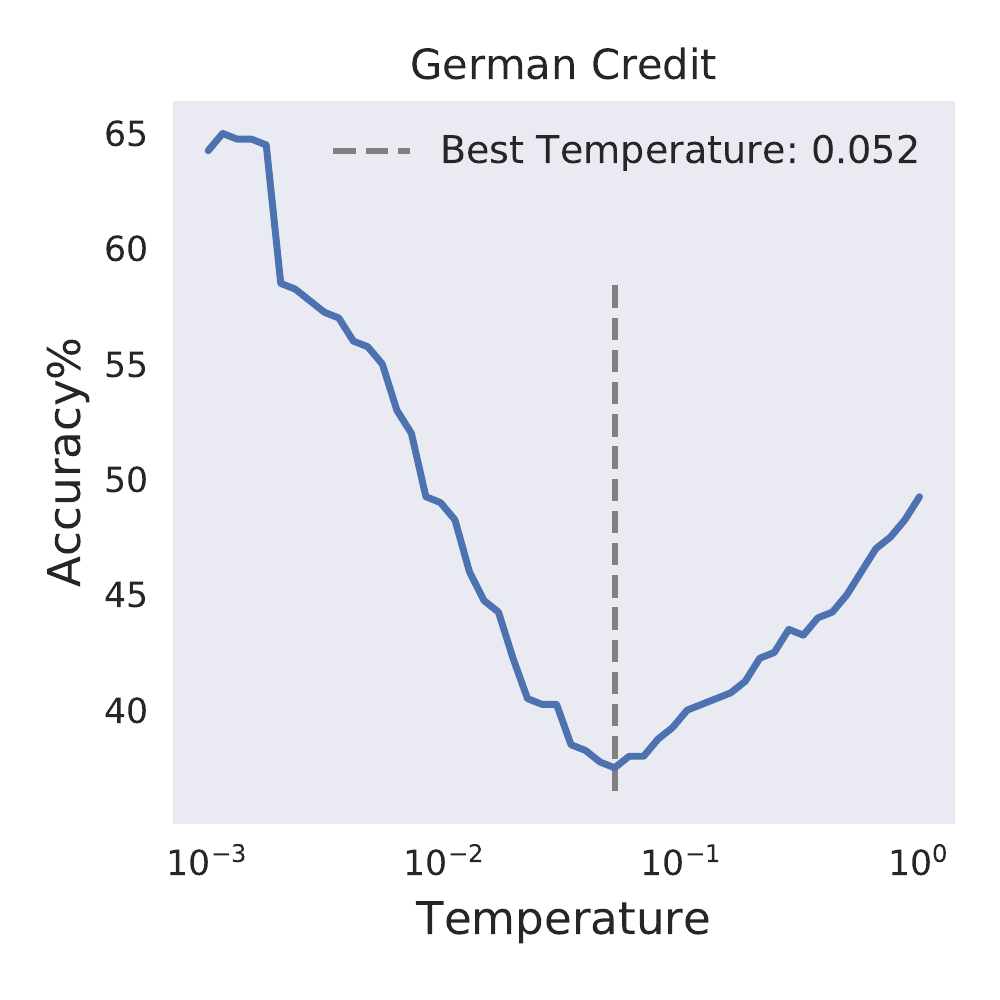}
         \caption{German Credit}
     \end{subfigure}
     \begin{subfigure}[b]{0.32\linewidth}
         \centering
         \includegraphics[width=0.9\linewidth]{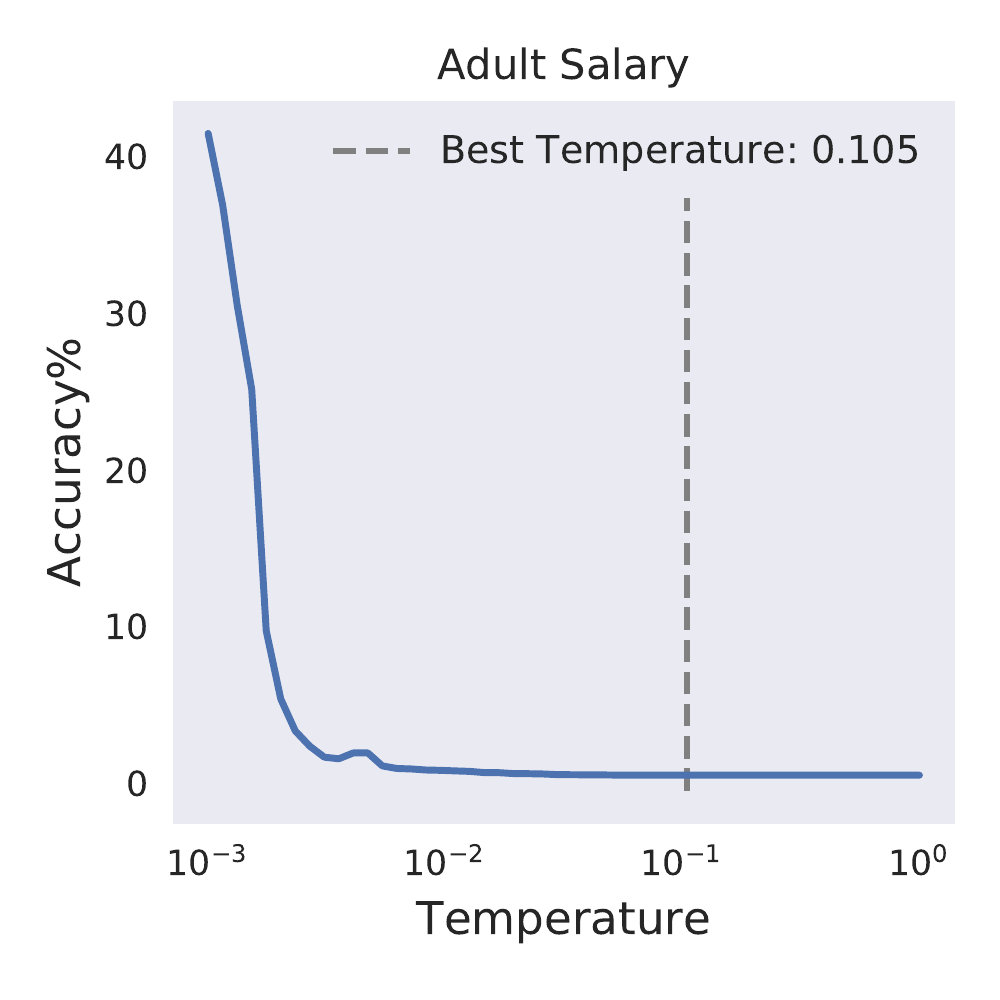}
         \caption{Adult Salary}
     \end{subfigure}
    %      \begin{subfigure}[b]{0.32\linewidth}
    %      \centering
    %      \includegraphics[width=1.0\linewidth]{figs/rf_digits_time.eps}
    %      \caption{Exhaustive}
    %  \end{subfigure}
     \begin{subfigure}[b]{0.32\linewidth}
         \centering
         \includegraphics[width=0.9\linewidth]{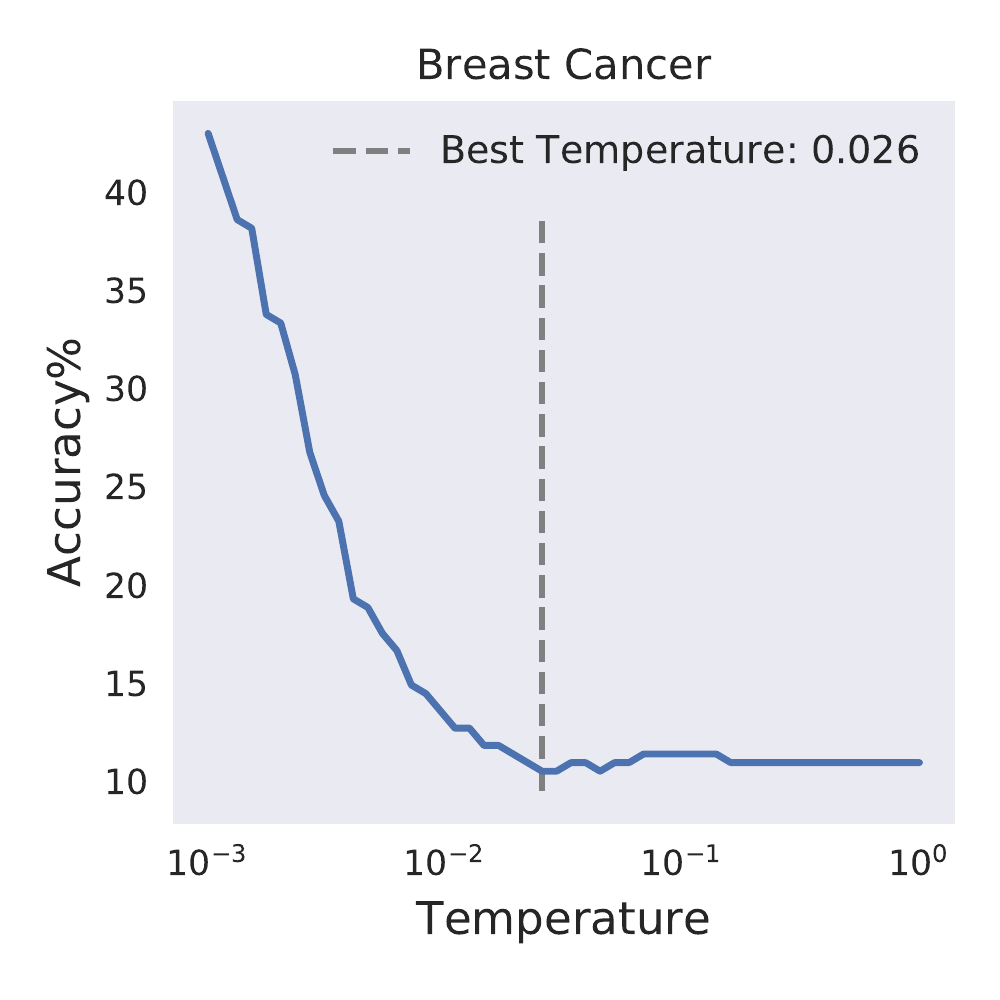}
         \caption{Breast Cancer}
     \end{subfigure}
    \caption{\textbf{Effects of Tuning Smoothing Temperature.} Temperature appears sensitive to the data (or application domain). Higher temperatures are not necessarily a good choice, as an adversarial example for the smoothed tree may not necessarily flip the decision of the original model.}
    \label{fig:hyper-parameter}
    \vspace{-5mm}
\end{figure}
\vspace{-2mm}
\subsection{Effectiveness of Tuning Hyper-Parameters}
\label{sec:hyper-parameter}
\vspace{-2mm}
As we demonstrated in Figure~\ref{fig:tempered_sigmoid}, the temperature hyper-parameter controls the smoothness of the smoothed model. In this experiment, we show how the temperature would impact the adversarial example search. As shown in Figure~\ref{fig:hyper-parameter}, with a relatively larger smoothing temperature ($10^{-1}$), the whitebox attack usually shows better performance than very low temperature ($10^{-3}$). This is because the low temperature preserves the original model's piece-wise property, which prevents gradient descent since the sigmoid functions are nearly saturated. However, this does not mean that a very high temperature is the best setting for all application domains. For the German Credit dataset (Figure~\ref{fig:hyper-parameter}(a)), we note that a large temperature ($10^{0}$) could result in poor performance. We note that the problem comes from the large approximation gap between the smoothed model and the original model. The adversarial examples of the smoothed model are no longer valid examples of the original model. Hence, tuning the temperature is a critical step to guarantee the performance of the proposed approaches. In this work, we tune the hyper-parameters through grid search in range of $[10e-3, 10e-2, 10e-1, 1]$  
\vspace{-2mm}
\subsection{Effectiveness of Noise Injection on Gradient Descent}
%fig:gradient_direction
\vspace{-2mm}
Figure~\ref{table:noise_vs_nonoise} shows the effect of adding noise during the adversarial example search. We note that the proposed approach (STA-Exhaustive) demonstrated remarkable performance improvement for two of three datasets in our experiment with noise injection. This observation reflects our intuition (as shown in Figure~\ref{fig:gradient_direction}) -- it is critical to add noise during the search on smoothed trees.  Since the smoothed tree ensembles still preserve some piece-wise property, pure gradient descent-based adversarial example search would either misguide the search direction or cause the search to get stuck from the beginning as demonstrated in Figure~\ref{fig:gradient_direction}.
%\chris{not sure I understand this last sentence. NOT having noise misguides the direction and could get stuck? or HAVING noise misguides the direction and could get stuck? Maybe reword the sentence?}

\begin{table}
\caption{\textbf{Performance comparison between pure gradient descent and noise injected gradient descent search.} All of the variants are based on STA-Exhaustive search. Lower is better. Statistic comes from 3-fold cross-validation.}
\centering
\resizebox{\textwidth}{!}{%
\begin{tabular}{c|c|c|c|c|c|c|c}
\toprule
\multirow{2}{*}{Data Name }     &   \multirow{2}{*}{Tolerance}   & \multicolumn{3}{c|}{Random Forest} & \multicolumn{3}{c}{XGboost} \\
\cline{3-8}
&&Original&Pure Gradient&Noise Injected&Original&Pure Gradient&Noise Injected\\
\midrule
 & $\epsilon$=0.2 & & $45.80 \pm 2.87 $ &  $44.30 \pm 2.89 $ & & $42.49 \pm 8.35$ &  $36.28 \pm 9.38 $ \\
    German Credit & $\epsilon$=0.5 &$71.89 \pm 1.83$ & $25.89 \pm 12.63$ &  $23.59 \pm 13.18$ &$75.29 \pm 3.13$ & $19.39 \pm 11.58$ &  $14.49 \pm 9.46$ \\
     & $\epsilon$=0.8 & & $22.89 \pm 11.21$ &  $19.97 \pm 13.08$ & & $14.99 \pm 8.90 $ &  $10.99 \pm 7.35$\\ \midrule
     & $\epsilon$=0.2 & & $7.09  \pm 0.15 $ &  $ 7.04 \pm 0.15 $ & & $0.87  \pm 0.31 $ &  $ 0.56 \pm 0.07$ \\
     Adult Salary  & $\epsilon$=0.5 & $77.59 \pm 0.27$ & $ 0.56 \pm 0.06 $ &  $0.43  \pm 0.04 $ &$85.28 \pm 1.68$ & $0.42 \pm 0.08  $ &  $ 0.31 \pm 0.08$\\
    & $\epsilon$=0.8 & & $ 0.20 \pm 0.11 $ &  $0.12  \pm 0.16 $ & & $0.41 \pm 0.10  $ &  $0.30  \pm 0.09$ \\ \midrule
      & $\epsilon$=0.2 & & $ 7.62 \pm 2.51 $ &  $6.21  \pm 2.39 $ & & $ 7.73 \pm 2.18 $ &  $7.56  \pm 2.23$\\
    Breast Cancer  & $\epsilon$=0.5 & $95.43 \pm 1.51$ & $ 0.20 \pm 0.21 $ &  $0.15  \pm 0.09 $ & $95.95 \pm 0.89$ & $ 0.17 \pm 0.25 $ &  $0.17  \pm 0.25$  \\
    & $\epsilon$=0.8 & & $ 0 \pm 0$ & $0 \pm 0$ & & $ 0 \pm 0$ & $0 \pm 0$\\
\bottomrule
\end{tabular}}
\label{table:noise_vs_nonoise}
\vspace{-5mm}
\end{table}

\vspace{-2mm}
\section{Conclusion and Discussion}
\label{sec:conclusion}
\vspace{-2mm}
This paper presents a novel adversarial robustness testing approach, STA, (with two variants) for tree ensemble models. The proposed method involves two steps. First, it smooths the tree ensemble to support auto-differentiation. Second, it conducts whitebox gradient-based attacks on the smoothed model for adversarial example search. To facilitate the adversarial example search, we introduce a number of techniques that yield remarkable performance improvement (in terms of effectiveness and efficiency), including temperature control, noise injection, feature-dependent perturbation bound, and log-derivative-based sampling. Our experiments in four application domains show that the proposed approach has a remarkable advantage over other state-of-the-art approaches in efficiency (more than ten times faster) while maintaining competitive effectiveness.

%{\color{red} The proposed approach is limited to adversarial robustness testing for tree ensemble models. It may not be suitable for other type of non-differentiable models. And, even though we introduced various tricks to maximize the coverage for the adversarial example search, our approach is not guaranteed to find all adversarial weakness of the model. It only serves as an fast and scalable checking approach.}

For as successful as our technique is, we emphasize we cannot guarantee it will find all model weaknesses. Adversarial examples present a distinct risk in using machine learning models outside of the lab, as their vulnerabilities may be exploited for malicious intents.  Accordingly, if our test incorrectly declares a model robust, there may be negative societal impact.  In addition, because our attack itself is powerful, it may be directly employed with malicious intent.  However, as our attack requires access to the model's internals, the likelihood of our test being misused in this way is reduced. 

\bibliographystyle{plain}
\bibliography{reference}
%\newpage
%\input{checklist}
\end{document}